%% file: main.tex
\documentclass{fairmeta}

\usepackage{template/macro}
\input{template/package}
\usepackage{xspace}

\newcommand{\AutoformBot}{\textsc{AutoformBot}\xspace}
\newcommand{\Atlas}{\textsc{ATLAS}\xspace}
\newcommand{\Mathlib}{\textsc{mathlib}\xspace}

\title{Formalizing Mathematics at Scale}

\author[1,2]{Ahmad Rammal}
\author[1,3]{Niket Patel}
\author[1,\dag,2]{Fabian Gloeckle}
\author[2,4]{Amaury Hayat}
\author[1,3]{Julia Kempe}
\author[1]{Remi Munos}
\author[1,*]{Charles Arnal}
\author[1,*]{Vivien Cabannes}

\affiliation[1]{FAIR at Meta}
\affiliation[2]{\small{CERMICS, ENPC, Institut Polytechnique de Paris}}
\affiliation[3]{\small{New York University}}
\affiliation[4]{\small{Korea Institute for Advanced Study}}

\contribution[*]{Equal contribution}
\contribution[\dag]{Work done while at Meta}

\abstract{\input{abstract}}

\date{\today}
\correspondence{Ahmad Rammal at \email{rammal@meta.com}}

\metadata[Code]{ \\ \url{https://github.com/facebookresearch/autoform-bot} \\ \space \url{https://github.com/facebookresearch/atlas-lean}}

\begin{document}

\maketitle

\input{core_arxiv}

\bibliographystyle{assets/plainnat}
\bibliography{references}

\beginappendix

\input{appendix}

\end{document}

%% file: abstract.tex
% Abstract text only — each entry point wraps in \begin{abstract}...\end{abstract}
% Title: ATLAS: Formalizing Mathematics at Scale
%As peer review syst ems in mathematics are increasingly strained, formal verification is becoming the only scalable path to rigor.
%Realizing this at the level of modern research mathematics requires a verified foundation far larger than today's MathLib, on the order of thousands of textbooks of prerequisite theory.
%

We present \AutoformBot, a multi-agent system for building an \textbf{A}utoformalized \textbf{T}extbook \textbf{L}ibrary \textbf{A}t \textbf{S}cale (\Atlas) in Lean 4.
\AutoformBot orchestrates thousands of LLM agents, equipped with formal verification tools, dependency-aware task scheduling, and collaborative version control, to translate informal textbook prose into machine-checked definitions and proofs.
We apply our methods to a corpus of \textbf{26} open-access textbooks spanning analysis, algebra, topology, combinatorics, and probability, producing \Atlas: a verified library of over \textbf{45{,}000} Lean~4 declarations and \textbf{500 thousand} lines of code.
We release two artifacts:
(i) \AutoformBot, the open-source multi-agent framework;
and (ii)~\Atlas, the resulting formal library.
Our results suggest that autoformalizing the core content of graduate-level mathematics at scale is now economically and technically feasible.
This opens the door to the automated verification of both human- and machine-generated mathematics at a research level. 

%% file: core_arxiv.tex
%% ============================================================================

\section{Introduction }
\label{sec:intro}

As large language models accelerate the production of ideas, code, and other artifacts, the \emph{verification} of their output becomes a major bottleneck.
In mathematics, this bottleneck is particularly acute: the peer review system is already under pressure, and referees increasingly lack the time to check every step of a proof, instead relying on their mathematical intuition to judge whether the proof \textit{appears} credible.
Once AI systems start generating mathematical reasoning at a pace that dwarfs human capacity, this trust-based verification model will become untenable.
The recent \emph{First Proof challenge}~\citep{abouzaid2026proof,armstrong2026takeaways} provided a glimpse of this future: LLMs rapidly generated dozens of solutions to challenging research questions, a few of which were correct and most of which were wrong in subtle, nontrivial ways.
Proof assistants offer a way out.

\paragraph{Proof assistants}
In a proof assistant, users express mathematical definitions, theorem statements, and proofs in a specialized programming language with a formal foundation, a process called \emph{formalization}; a small engine (the \emph{kernel}) checks every logical step mechanically, so that any accepted proof is guaranteed to be valid, relative to the stated definitions and assumptions.
Thus, verifying a formalized proof reduces to checking that it is accepted by the proof assistant and that the formal statement itself faithfully captures the intended meaning.
A simple example of a formalized statement is given in Figure~\ref{fig:formalization_example}.
We use Lean~4~\citep{moura2021lean4}, a modern proof assistant that doubles as a general-purpose programming language, as our formalization environment. 
Other widely used proof assistants include, for instance, Rocq/Coq~\citep{bertot2004interactive},
Isabelle/HOL~\citep{nipkow2002isabelle}, HOL Light~\citep{harrison2009hol},
Agda~\citep{bove2009brief}, Mizar~\citep{grabowski2010mizar,bancerek2018mizar},
and Metamath~\citep{megill2019metamath}.

\begin{figure}[t]
  \centering
  \begin{tcolorbox}[
    enhanced, arc=6pt, boxrule=0.5pt,
    colback=yellow!12, colframe=yellow!55!black,
    width=0.495\linewidth, before=, after=\hfill,
    left=5pt, right=5pt, top=2pt, bottom=2pt,
  ]
    { For all $x \in \mathbb{N}$, if $x < 2$, then $x + 3 < 5$.
    \tcbline
    Let $x < 2$. Adding $3$ to both sides preserves the inequality, so $x + 3 < 2 + 3 = 5$.
    }
  \end{tcolorbox}%
  \begin{tcolorbox}[
    enhanced, arc=6pt, boxrule=0.5pt,
    colback=brown!15, colframe=brown!60!black,
    width=0.495\linewidth, before=, after=,
    left=5pt, right=5pt, top=2pt, bottom=2pt,
  ]
    {\ttfamily
      example (x : Nat) (h : x < 2) :
      x + 3 < 5 :=
    \tcbline
      by\\
      \hspace*{1em}exact Nat.add\_lt\_add\_right h 3
    }
  \end{tcolorbox}
  \caption{A simple mathematical statement and proof, shown informally and as a Lean formalization.}
  \label{fig:formalization_example}
  \end{figure}

\paragraph{Laying out foundations}
Formalizing mathematics is cumulative: before one can state a theorem about, say, compact metric spaces, one needs formal definitions of metrics, topological spaces, compactness, and the lemmas connecting them.
\Mathlib~\citep{mathlib2020}, Lean's community-maintained mathematical library, is an ongoing effort to build such foundations.
 With roughly 2.1 million lines of code and years of expert effort behind it, \Mathlib{} covers broad but incomplete swaths of the mathematical landscape; while some domains, such as algebra or category theory, are well-represented, large gaps remain in others, such as differential geometry or PDEs.
As a result, most current mathematical research cannot be formalized in Lean without prohibitively time-consuming preparatory work.

Textbooks are a natural unit of work for closing this gap: by their very nature, they provide solid and general foundations upon which to build. While formalizing an entire textbook is a tremendous undertaking for a human expert, recent progress in the capabilities of frontier models has opened the door to the automated formalization, or \textit{autoformalization}, of large corpora of texts~\citep{wu2022autoformalization,urban2026130klinesformaltopology,wang2026m2fautomatedformalizationmathematical,gloeckle2026automatic}.

\paragraph{Benefits of autoformalization at scale}
Efficient automated formalization would make large-scale LLM-powered mathematical research possible: large groups of human experts and LLM agents could collaborate in a process reminiscent of open-source software, working independently on modular, well-specified pieces while the proof assistant ensures that their contributions compose correctly. Without such mechanical verification, the resulting body of work would exceed any individual's capacity to check.
It would additionally allow for the validation of existing mathematical knowledge, which has been one of the main goals of formalization efforts so far.
Finally, autoformalization could be used to produce trustworthy reward signals for mathematical reasoning trajectories when training LLMs with reinforcement learning. Indeed, current methods, which involve training on problems with explicit and easily verifiable solutions (``$2$'', ``$\sqrt{x+1}$'', \ldots) or relying on other LLMs as judges~\citep{cobbe2021gsm8k,hendrycks2021math,zheng2023judging}, might not scale to the research-level mathematics that frontier models are starting to tackle~\citep{feng2026aletheia}.

\paragraph{Multi-agent scaffolds}
While frontier models have made considerable progress in Lean in recent years, they still trail behind in performance compared to other programming languages; Claude Opus 4.6, Gemini 3.1 Pro, or GPT 5.4~\citep{anthropic2026claude46,google2026gemini31pro,openai2026gpt54} typically fail in their initial attempt to provide a 20-line proof to a standard mathematical statement, but eventually succeed when given enough time and tools. As such, the formalization of an average textbook is a codebase-building endeavor far beyond what LLMs can achieve in a single, unsupervised shot.
This creates the need for adapted scaffolds that provide tools, frameworks, and supervision and can leverage parallelism by coordinating the work of dozens or hundreds of agents.

\paragraph{Contributions}
In this work, we introduce \AutoformBot, a multi-agent framework for formalizing mathematical textbooks at scale, whose agents are powered through user-supplied access to frontier-model APIs.
Scaling from a single coding agent to thousands working on a shared repository raises hard coordination problems: agents make incompatible design decisions, generate duplicate work, pursue tangential goals, and create cascading failures through a shared merge queue.
\AutoformBot treats formalization as a collaborative software engineering problem, coordinating agents through mechanisms that are well represented in their training data: git branches, pull request review, and issue tracking.
As formal verification provides a sharp coordination signal, textbook formalization becomes an ideal testbed for multi-agent orchestration research.

We release two artifacts:
\begin{enumerate}
  \item \textbf{\AutoformBot: an open-source multi-agent framework}
  for textbook autoformalization, featuring formal verification tools, dependency-aware task scheduling, and collaborative version control. It includes a visualizer for human-in-the-loop interaction, a rigorous evaluation pipeline, and can be easily
  connected to user-supplied models, whether accessed through APIs or hosted locally.

  \item \textbf{\Atlas: verified formal libraries at scale.}
  We apply \AutoformBot, powered primarily by Opus 4.6, to \textbf{26} open-access mathematical textbooks and produce verified Lean~4 libraries totaling over \textbf{45{,}000} declarations spanning several areas of mathematics and extending \Mathlib{}'s coverage.
  We intend to keep improving this initial effort until it becomes complete and systematic enough to serve as a seamless machine-generated extension of \Mathlib{}.
\end{enumerate}

Beyond the artifacts, we report empirical observations on multi-agent coordination at scale: an adversarial dynamic between workers and reviewers around verification circumvention, context degradation in long-lived agents that motivates delegation to specialized short-lived ones, and ablations quantifying the contribution of each feedback component.

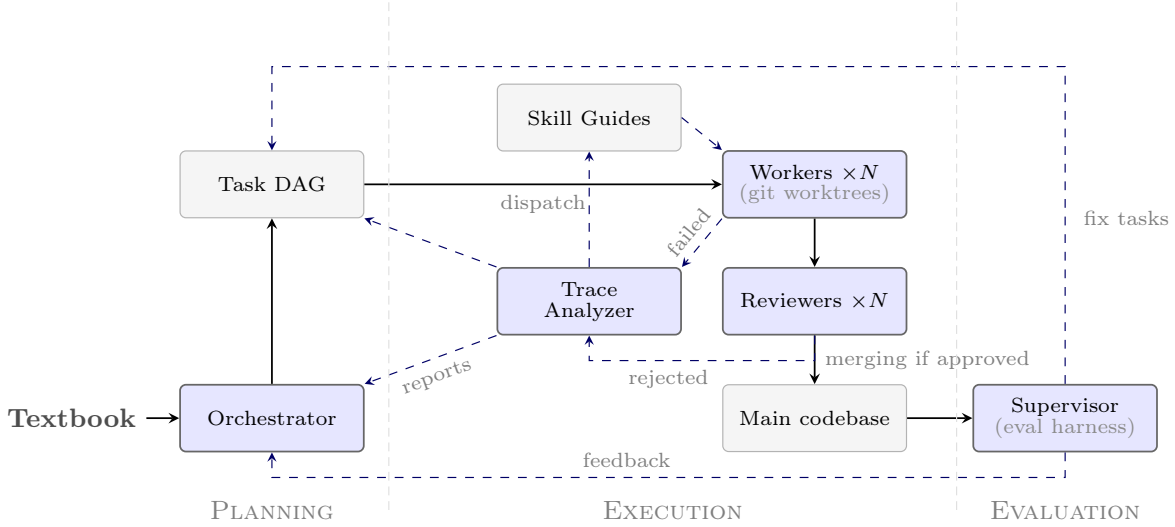
\begin{figure}[t]
  \centering
  \resizebox{.95\columnwidth}{!}{\input{figures/pipeline}}
  \caption{%
    \AutoformBot architecture.
    The orchestrator reads the book and builds a task DAG.
    Workers formalize individual statements in isolated worktrees; successful builds pass through concurrent review and a batched merge queue.
    The trace analyzer learns from task failures and writes skill guides for subsequent attempts.
    The supervisor evaluates book targets after each merge and dispatches fix tasks via triage agents.
    Arrows indicate information flow; dashed arrows indicate feedback loops.
  }
  \label{fig:pipeline}
\end{figure}

%% ============================================================================
\section{Related Work}
\label{sec:related}
%% ============================================================================

\paragraph{LLM-based theorem proving and autoformalization.}
A growing line of work applies language models to formal proof generation. 
GPT-$f$~\citep{polu2020generative,polu2022formal} first demonstrated that transformers can generate useful proof steps in formal languages. \cite{lample2022hypertree} further showed that combining transformers with a reinforcement learning approach could strongly increase performance.
Subsequent systems such as LEGO-Prover~\citep{wang2024legoprover}, AlphaProof~\citep{hubert2025alphaproof}, ABEL~\citep{gloeckle2024abel} or more recently DeepSeek-Prover~\citep{xin2024deepseekprover,deepseekproverv2}, Goedel-Prover~\citep{lin2025goedelprover}, Kimina-Prover~\citep{wang2025kiminaprover}, Seed-Prover~\citep{seedprover2025} have been developed and pushed the state of the art on benchmarks like miniF2F~\citep{zheng2022minif2f}, ProofNet~\citep{azerbayev2023proofnet}, the International Mathematical Olympiads, or Putnam Bench \citep{tsoukalas2024putnambench}. 
Other systems such as \citet{achim2025aristotle} and \citet{chen2026fel} have been able to provide formal proofs to research-level questions.

On the autoformalization side, since \citet{wu2022autoformalization}, a large body of work has studied autoformalization datasets 
\citep[e.g.][]{azerbayev2023proofnet,ying2024lean,ju2026ai,liu2026numina}. 
Among the milestones, one can cite, for instance, large-scale formalizations of competition problems \citep{hubert2025alphaproof,numina2025lean} or, recent hybrid approaches between manual and automatic formalization leading to the formalization of the strong prime number theorem \citep{mathinc_strongpnt}, the sphere packing theorem \citep{hariharan2026spherepacking} or De Giorgi-Nash-Moser theory \citep{armstrong2026formalizationgiorginashmosertheorylean}. These approaches target isolated theorems; none attempts the systematic formalization of a full textbook.

\paragraph{Large-scale textbook formalization.}
Several recent efforts have begun to tackle longer documents.
\citet{urban2026130klinesformaltopology} used a single coding agent to formalize initial chapters of a point-set topology textbook, in Megalodon.
\citet{wang2026m2fautomatedformalizationmathematical} presented a two-phase multi-agent scaffold formalizing an introductory analysis textbook of 300 pages producing 40k lines of Lean, though with a significant \Mathlib{} overlap. \citet{gloeckle2026automatic} provided the first example of graduate textbook fully autoformalized without human intervention in one week.
% \todo[author=Amaury]{I'd suggest to be nicer / to acknowledge more / less reduce the contribution of M2F (e.g. introductory analysis textbook of 300 pages producing 40k lines of Lean with, however, a significant \Mathlib{} overlap): we have nothing to hide and otherwise reviewers (who will very likely know their work) may believe that we try to diminish their work because this work is not so different (while it is, there is nothing to hide). I have the same remark for MathInc and Aristotle}.
% \todo[author=Amaury]{MathInc and Aristotle released work is not about textbook formalization so it may be weird to have them in this paragraph, I would suggest to move them to the previous one (they're still isolated theorems, even with large dependencies they precisely don't do textbook). Also, I think one should acknowledge more (even in just one sentence) what they did (e.g. formalized landmark research results and cite their papers). A minor form remarks: if ``Math, Inc.'s Gauss'' then one needs to put ``Harmonic's Aristotle''.}.
Our work differs from these works both in scale and methodology: we formalize \textbf{26} books across diverse mathematical areas, using an open-source framework with detailed compute cost and efficiency reporting.

\paragraph{Multi-agent systems for software engineering.} %\charles{shorten this paragraph}
SWE-agent~\citep{yang2024sweagent} demonstrated effective single-agent software engineering, while MetaGPT~\citep{hong2024metagpt} introduced role-based multi-agent coordination with structured communication.

\section{Success Criteria and Evaluation Harness}
\label{sec:eval-harness}

Before describing the pipeline, we define what constitutes a successful formalization and the harness used to measure it.
In Lean, one can avoid providing a proof by declaring it an axiom or by replacing the proof with the \texttt{sorry} keyword.
Our policy is that every statement whose proof appears in the source material must be properly formalized, without \texttt{sorry} or axioms; statements whose proof is not provided (e.g.\ because the author defers to a reference) may be axiomatized.
However, an autoformalization attempt can fail in several ways.
It may fail by illegitimately using an axiom or \texttt{sorry}.
More subtly, a statement may appear fully proved but depend on another declaration whose proof contains an axiom that propagates silently through the dependency chain. Finally, an agent may produce structurally degenerate formalizations that technically compile but do not faithfully capture the underlying mathematics (e.g.\ a theorem proved only for $\mathbb{F}_p$ instead of for all groups, or a definition that bakes in what should be a separate theorem).

We start each attempt by running a preprocessing script that identifies all formalizable statements in the source textbook; we call these \textit{target statements} or \textit{goals}.
A statement is successfully formalized if it faithfully captures the mathematical content and its proof does not directly rely on any illegitimate axiom or \texttt{sorry}. This criterion is non-transitive: if a correct proof invokes a target lemma whose own proof contains a \texttt{sorry}, the calling statement is deemed successful but the lemma is not.
Our main success metric is the number of successfully formalized target statements.

\begin{figure}[t]
  \centering
  \resizebox{.7\columnwidth}{!}{\input{figures/dependency_cone_II}}
  \caption{%
    Dependency cone of a target declaration. Edges represent direct
    dependencies between project-local declarations. Structural tags
    and gap markers propagate upward as alerts on the target.
  }
  \label{fig:dep-graph}
\end{figure}
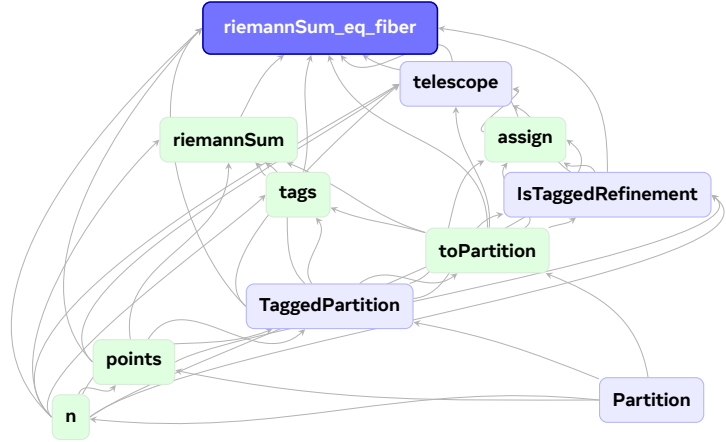

Detecting illegitimate axiomatizations requires recursively checking the declarations a statement depends on, rather than inspecting individual declarations in isolation.
A direct \texttt{sorry} is easy to spot, but an axiom or weakened definition that propagates through a chain of helper lemmas can only be caught by analyzing the dependency structure of the entire project.
To this end, we build a \emph{declaration dependency graph} by running a Lean metaprogram inside the compiled project (Figure~\ref{fig:dep-graph}).
The metaprogram walks every project-local declaration and extracts its nature (theorem, axiom, etc.), the set of project-local declarations it references, and its axiom set.
For each declaration, we also compute \emph{structural tags} by inspecting the proof term, flagging patterns that may indicate faithfulness defects: vacuous bodies, hypothesis smuggling, trivially constructed instances, and others.

The evaluation harness uses the dependency graph together with a set of mechanical and LLM-based checks, applied in three stages:
\begin{enumerate} [leftmargin=*]
  \item \textbf{Mechanical gates.}
    The project must compile without errors, and source files must not contain metaprogramming keywords (\texttt{elab},
    \texttt{syntax}) that could make the code semantically misleading.
  \item \textbf{Matching.}
    For each target statement, a matcher agent finds its corresponding Lean declaration in the generated codebase.
  \item \textbf{Statement-level grading.}
    Three independent LLM judges score each matched target along the rubrics described below.
\end{enumerate}
The dependency graph is exposed to the judges as a queryable tool, allowing them to trace sorry chains, inspect structural tags, and investigate suspicious dependencies when scoring.
The three rubrics are:
\begin{enumerate}[leftmargin=*]
  \item \textbf{Faithfulness:}
    whether the Lean statement captures the source material's mathematical content (hypotheses and conclusions) at full strength, without weakening the result or hiding content in typeclass fields.
  \item \textbf{Proof integrity:}
    whether the proof chain represents genuine mathematical work---no unjustified \texttt{sorry} or axioms, no orphan classes used as hypotheses, no vacuous definitions.
  \item \textbf{Code quality:}
    adherence to Mathlib conventions~\citep[using intelligence collected by][]{mathlibconventions}---naming, tactic choice, typeclass generality, and proof structure.
\end{enumerate}
Our evaluation harness is released as a self-contained module in the accompanying codebase.

\section{\AutoformBot}
\label{sec:method}
%% ============================================================================

\AutoformBot casts textbook formalization as a collaborative software engineering problem.
From a software engineering perspective, a formalization project is simply a code repository in a special-purpose programming language (Lean~4) that must build without errors and pass quality checks regarding faithfulness to the source material.
Coding agents---persistent LLM instances equipped with code-related tools---fit this task well: they can read files, make edits, run terminal commands, and interact with the Lean type checker.
The challenge is scaling from one agent to thousands while maintaining coherence.
\AutoformBot is an agentic harness meant to address this challenge by equipping individual agents---powered through access to any served model---with tools geared towards formal verification capabilities, a shared infrastructure for task tracking and version control, and a flexible collaboration architecture.

\subsection{Pipeline Architecture}
\label{sec:orchestrator-pipeline}

The pipeline is a multi-agent system organized in three management tiers (Figure~\ref{fig:pipeline}): a high-level \emph{orchestrator} that plans work from the book's mathematical structure, mid-level agents---the \emph{trace analyzer} and the \emph{supervisor}---that manage learning at the task level and evaluation at the target level, and low-level \emph{workers} and \emph{reviewers} that execute individual formalization tasks.
These tiers communicate through a shared coordination infrastructure described further below in \S\ref{sec:infrastructure}.
Each agent is powered through access to the endpoint of a served model---typically Opus 4.6 in our experiments.
This architecture sets it apart from the earlier, more loosely-organized system in \citet{gloeckle2026automatic}.
% This sophisticated architecture sets it apart from earlier, more loosely-organized systems such as \cite{gloeckle2026automatic}.

\paragraph{Planning: the orchestrator.}
The orchestrator is a long-lived LLM agent that reads the source textbook and encodes the work to be done as a task Directed Acyclic Graph (DAG), whose nodes are formalization targets (definition, theorem, etc.) and whose edges encode the book's logical dependencies---if theorem~B uses definition~A, then the task for~B depends on the task for~A.
It continuously updates the DAG as the project evolves, and maintains a persistent TODO list on disk to track patterns and issues across rounds, compensating for the limited context window of a long-running conversation.

\paragraph{Execution: workers and reviewers.}
Tasks are dispatched by a runner that continuously polls for ready tasks (all dependencies met) and assigns them to available workers.
A worker's output must pass a series of quality checks (see Section~\ref{sec:eval-harness}), including inspection by a \textit{reviewer}.
Optionally, multiple workers can race on the same target, each in its own worktree; the first to clear all quality gates wins and the remaining attempts are cancelled.
Approved changes enter a merge queue (see \S\ref{sec:infrastructure}).

\paragraph{Task-level feedback: the trace analyzer.}
When a task fails, the system must learn from the failure so that the
next attempt does not repeat the same mistakes.
The \emph{trace analyzer} is a persistent agent assigned to each failed task.
It maintains \emph{skill guides} containing lessons from past attempts; workers are required to read this guide before starting a new attempt. It acts on the DAG based on feedback from the workers about task difficulty and potential decompositions, and generates a structured report summarizing the failure, which the orchestrator consumes during its next planning
round.

\paragraph{Target-level feedback: the supervisor.}
The \emph{supervisor} loop operates at the level of the list of targets, driven by a \emph{goal tracker} that records the status of each target statement (pending, completed, or failed) independently of the task DAG.
After each successful merge, the supervisor computes the git diff, uses a matcher agent to identify which target statements were affected, and runs the evaluation harness (Section~\ref{sec:eval-harness}) on the affected targets in an isolated worktree.

For targets that fail evaluation, a \emph{triage agent} creates granular fix tasks in the DAG (e.g., one task per unjustified \texttt{sorry}).
Both the trace analyzer and the triage agents were motivated by the same observation: delegating targeted analysis to fresh, task-scoped agents produces better results than burdening the long-lived orchestrator, whose response quality degrades over dozens of planning rounds as its context grows---a phenomenon we refer to as \emph{LLM fatigue}.

\subsection{Tool design}

\AutoformBot equips its agents with tools.
Tool servers expose capabilities via the Model Context Protocol~\citep[MCP,][]{mcp2024}; the framework converts MCP tool schemas into native function-call definitions for the underlying LLM, so that models interact with tools through their standard function-calling interface.
Agents are equipped with a configurable set of MCP tool servers, organized by category:

\begin{itemize}[nosep,leftmargin=*]
  \item \textbf{Execution.} A Lean REPL~\citep{morrison_repl} for running Lean code interactively, and a Lean LSP server~\citep{dressler2024leanlspmcp} for in-file diagnostics and proof-state queries.
  \item \textbf{Filesystem and search.} Sandboxed filesystem access, grep search, and a \Mathlib{} search index powered by Loogle~\citep{loogle} for type-based declaration search.
  \item \textbf{Version control.} Git operations and worktree management (creation, synchronization, cleanup), with shared \Mathlib{} installation via symlinks to keep worktrees lightweight.
  \item \textbf{Orchestration.} Sub-agent spawning, task dispatch, trackers for tasks and targets, and issues with dependency-aware status lifecycle, scratchpad for persistent notes, job scheduling, and trace inspection for debugging completed runs.
  \item \textbf{Communication.} User-to-agent and agent-to-user communication.
  \item \textbf{Discovery.} Skill loading from accumulated knowledge.
\end{itemize}
The exact set of tools and permissions granted to an agent depends on its role.

\subsection{Coordination Framework}
\label{sec:infrastructure}

Agents exist within an infrastructure framework that conditions their actions and with which they can interact through their tools.

\begin{itemize}[nosep,leftmargin=*]
\item \textbf{Task tracker.}
The persistent DAG of tasks with explicit dependency edges and a fixed status lifecycle (pending $\to$ in-progress $\to$ completed, failed or deleted, ready to be dispatched or not).

\item \textbf{Git worktree isolation.}
Every agent operates in a short-lived \texttt{git worktree} branching from the shared repository.
Completed work is integrated via rebase-then-fast-forward merge, preserving a linear history on the main branch.

\item \textbf{Concurrent racing.}
Several worker agents can race on the same task, each working in its own worktree. 
The first agent to clear all gates wins.

\item \textbf{Resource budgeting.}
A \emph{resource pool} wraps asyncio semaphores with active-count tracking to cap concurrent LLM calls and tool invocations.

\item \textbf{Process management.}
Agents rely on long-lived subprocesses (Lean REPL sessions, inference servers) that require careful lifecycle management.
A thread-based session pool provides queue-based load balancing across stateful RPC backends, background monitors enforce per-process memory limits, and a graceful teardown routine cleans up entire process trees on agent completion or failure.

\item \textbf{Multi-node execution.}
The executor abstraction that mediates between the scheduling loop and agent pools has both a local and a distributed implementation.

\item \textbf{Merge queue.}
We use a batched merge queue inspired by~\citet{borsng}: pending merges produced by workers are collected, rebased sequentially onto \texttt{main}, and verified with a single build over the combined result.
If it fails, the queue bisects the batch to isolate the breaking commit, lands the good prefix, and rejects the offender.

\item \textbf{Escalation protocol.}
Workers can file structured escalations that are consumed by the trace analyzer and surfaced to the user (see \S\ref{subsec:human_in_the_loop}).
\end{itemize}

\subsection{Human-in-the-loop and visualization}
\label{subsec:human_in_the_loop}

\begin{figure*}[t]
  \centering
  \includegraphics[width=\textwidth]{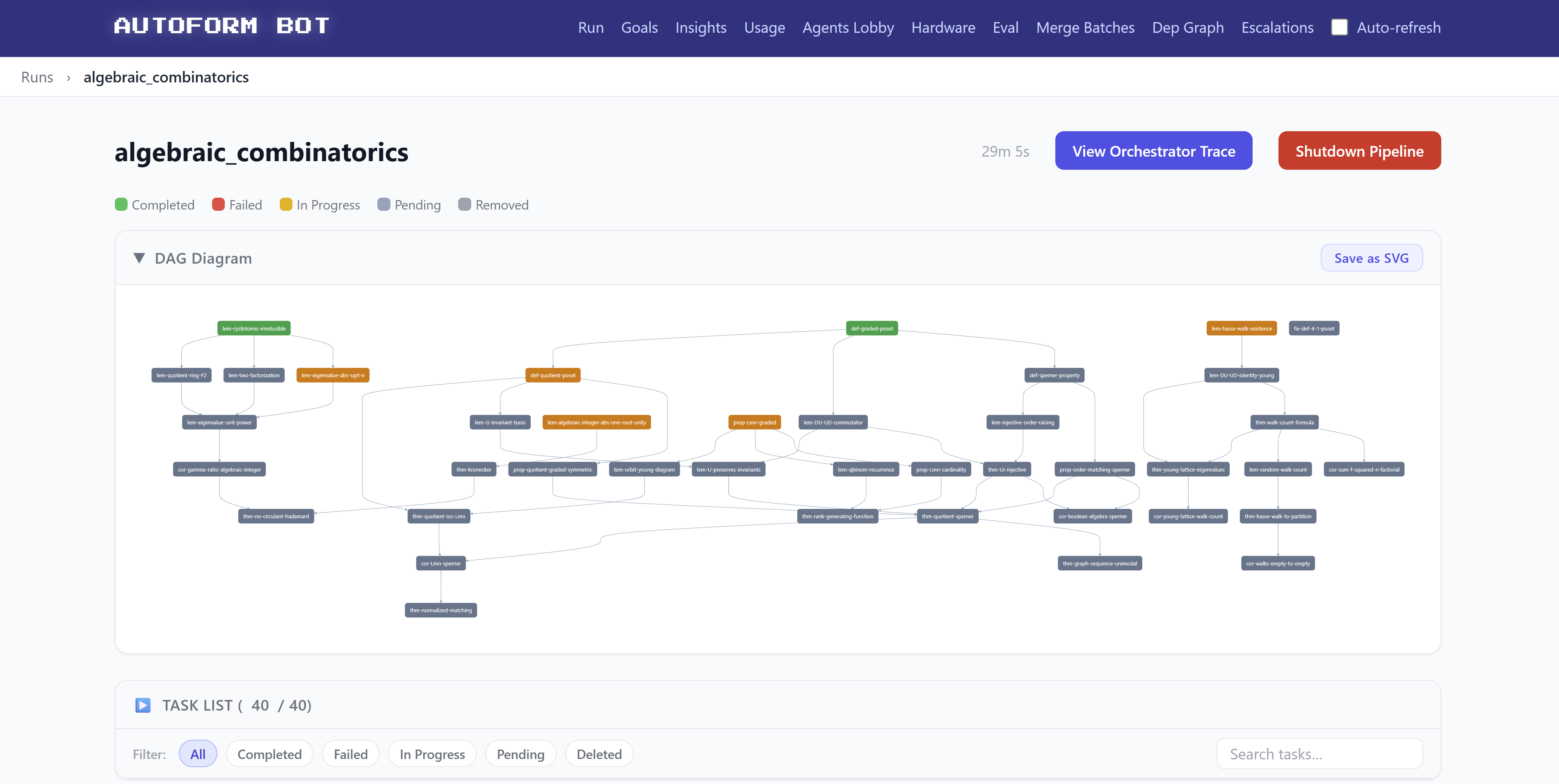}
  \caption{The graph of tasks of a formalization attempt, as shown by \AutoformBot's visual interface.
  }
  \label{fig:visualizer}
\end{figure*}

Though \AutoformBot is designed to operate fully independently, we also provide a convenient visual interface for tracking key metrics (compute spent, number of formalized statements, flagged issues, infrastructure usage, \ldots), inspecting the graph of logical dependencies, and enabling bidirectional communication with agents through escalation messages and user directives.
A screenshot of one of the tabs of the visual interface is shown in Figure~\ref{fig:visualizer}.

\label{sec:fortlib}\begin{table}[h]
  \caption{Per-book formalization results.
  For each book we report formalized target statements out of the total identified in the source, lines of Lean~4 code (excluding comments and blank lines), and tokens consumed (Appendix~\ref{app:compute_measure}).}
  \label{tab:all-books}
  \centering
  \scriptsize
  \begin{tabular}{llrrr}
    \toprule
    Course & Area & Formalized Statements & LoC & Tokens (M) \\
    \midrule
    \emph{Algebra Notes I \& II} & Algebra & 151/176 (85.8\%) & 4{,}409 & 1{,}963 \\
    \emph{Algebraic Combinatorics} & Combinatorics & 37/39 (94.9\%) & 9{,}343 & 1{,}441 \\
    \emph{Algebraic Geometry I} & Alg.\ Geometry & 112/186 (60.2\%) & 27{,}393 & 7{,}629 \\
    \emph{Algebraic Topology I} & Topology & 110/171 (64.3\%) & 20{,}143 & 10{,}323 \\
    \emph{An Algorithmist's Toolkit} & Combinatorics & 131/158 (82.9\%) & 8{,}234 & 2{,}004 \\
    \emph{Arithmetic Geometry} & Number Theory & 266/335 (79.4\%) & 29{,}573 & 11{,}101 \\
    \emph{Boolean Functions} & Combinatorics & 44/108 (40.7\%) & 7{,}949 & 2{,}327 \\
    \emph{Buildings} & Algebra & 44/74 (59.5\%) & 48{,}809 & 20{,}443 \\
    \emph{Combinatorial Optimization} & Combinatorics & 22/36 (61.1\%) & 7{,}934 & 2{,}476 \\
    \emph{Complex Variables} & Analysis & 37/38 (97.4\%) & 6{,}225 & 1{,}251 \\
    \emph{Differential Analysis} & Analysis & 88/113 (77.9\%) & 23{,}713 & 11{,}743 \\
    \emph{Differential Geometry} & Geometry & 112/147 (76.2\%) & 8{,}942 & 1{,}934 \\
    \emph{Elliptic Curves} & Number Theory & 212/360 (58.9\%) & 22{,}316 & 11{,}058 \\
    \emph{Fourier Analysis} & Analysis & 34/38 (89.5\%) & 6{,}671 & 1{,}186 \\
    \emph{Geometry of Manifolds} & Geometry & 40/72 (55.6\%) & 16{,}408 & 6{,}865 \\
    \emph{High Dimensional Statistics} & Probability \& Statistics & 65/73 (89.0\%) & 31{,}715 & 975 \\
    \emph{Intro.\ to Functional Analysis} & Analysis & 68/72 (94.4\%) & 2{,}006 & 554 \\
    \emph{Intro.\ to PDEs} & PDEs & 86/105 (81.9\%) & 20{,}740 & 2{,}972 \\
    \emph{Lie Groups} & Algebra & 74/185 (40.0\%) & 50{,}594 & 45{,}384 \\
    \emph{Number Theory I} & Number Theory & 460/576 (79.9\%) & 54{,}760 & 15{,}424 \\
    \emph{Probabilistic Methods in Combinatorics} & Combinatorics & 109/210 (51.9\%) & 15{,}604 & 2{,}720 \\
    \emph{Projection Theory} & Analysis & 73/111 (65.8\%) & 9{,}672 & 2{,}678 \\
    \emph{Real Analysis} & Analysis & 175/177 (98.9\%) & 2{,}224 & 586 \\
    \emph{Tensor Categories} & Algebra & 137/229 (59.8\%) & 29{,}729 & 11{,}338 \\
    \emph{Theory of Computation} & Computer Science & 84/118 (71.2\%) & 10{,}581 & 3{,}580 \\
    \emph{Theory of Probability} & Probability \& Statistics & 84/100 (84.0\%) & 8{,}231 & 3{,}201 \\
    \midrule
    \textbf{Total (26 books)} & & \textbf{2{,}855/4{,}007 (71.3\%)} & \textbf{483{,}918} & \textbf{183{,}157} \\
    \bottomrule
  \end{tabular}
\end{table}

\section{Results and Analysis}
\label{sec:results}

\subsection{\Atlas: verified formal libraries}

We have applied \AutoformBot powered by Opus 4.6 to a corpus of \textbf{26} open-access mathematical textbooks spanning analysis (real, complex, functional, Fourier, and differential), algebra (abstract and Lie theory), algebraic and differential geometry, algebraic topology, number theory (algebraic, arithmetic, and elliptic curves), combinatorics, partial differential equations, probability and statistics, and theoretical computer science.
Human involvement was minimal, ranging from nonexistent to sending a few messages of general advice ("Try decomposing this difficult task into helper lemmas") once a day depending on the book. The pipeline ran for about a week on each book, though increased parallelism can speed up the process.

The resulting library comprises over \textbf{45{,}000} verified Lean~4 declarations and about \textbf{500 thousand} lines of code, with the precise decomposition detailed in Table~\ref{tab:all-books}. This is comparable in order of magnitude to \Mathlib{}'s 2.1 million lines of code and $308{,}129$ declarations~\citep{li2026network}.
Each book gives rise to a self-contained Lean project that depends on \Mathlib{} and builds without errors.
Definitions and theorem statements are checked for faithfulness to the source material through the evaluation harness (\S\ref{sec:eval-harness}).
We release \Atlas as open-source repositories, one per book, with full provenance linking each formal statement back to its source text.

\begin{figure*}[t]
  \centering
  \includegraphics[width=0.9\textwidth]{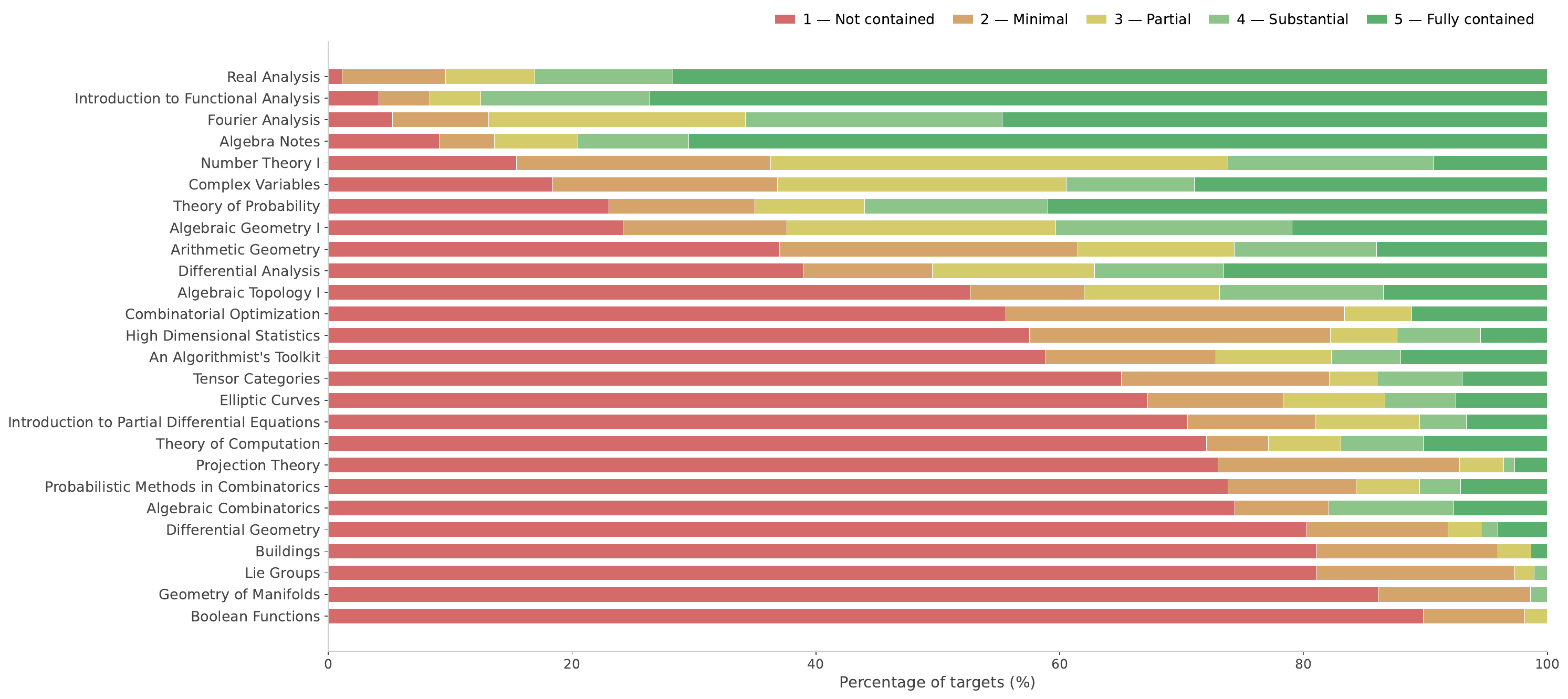}
  \caption{For each statement within each book, we estimate its difficulty in terms of how much infrastructure required to formalize it is missing from \Mathlib{}, and report these statistics for each book. The rubric used for this estimation is provided in Appendix~\ref{app:containment_rubric}.}
  \label{fig:difficulty}
\end{figure*}

As can be observed in Table~\ref{tab:all-books}, none of the books are fully formalized.  Most books contain a few particularly challenging statements, often because the required mathematical infrastructure is both absent from \Mathlib{} and not developed in detail in the book itself; such statements would present similar challenges to human annotators.
Our policy was to stop formalization processes once they reached a phase of strongly diminishing returns, where each new success required exponentially more compute, in order to allocate our resources to the coverage of more books.
In Figure~\ref{fig:difficulty}, we report an estimate of the difficulty of each statement in each book, in terms of how much infrastructure required to formalize it is missing from \Mathlib{}.

Compute costs are dominated by the workers (see Appendix~\ref{app:compute_measure} for a precise decomposition).
Though exact numbers depend on the chosen API and the provider's pricing, we estimate that our pipeline is already cheaper per line of code than expert human annotators, while being faster and more scalable.
%that the monetary cost per line of code of running our pipeline is 2 to 4 times lower than that of expert human annotators (paid around \$100 an hour), while being faster and more scalable.
On the other hand, the overall quality of the output remains inferior to that of expert-written Lean code (as flagged by both our harness and visual inspection by human experts).

%In what follows, we run our ablations on a comparatively easily formalizable book, \textit{Algebraic Combinatorics} by Richard Stanley.

\paragraph{Trustworthiness of the evaluation harness} Our evaluation harness relies on a combination of mechanical checks and LLMs as judges, which should not a priori be fully trusted. However, direct inspection by human experts (see Appendix~\ref{app:annotators_feedback}) of the formalization of our example book aligns with the conclusions of our harness, which strengthens our confidence in its judgment.

%\paragraph{Ablations.}
\subsection{Ablations}
\label{sec:ablations}

\begin{figure*}[t]
  \centering
  \includegraphics[width=\textwidth]{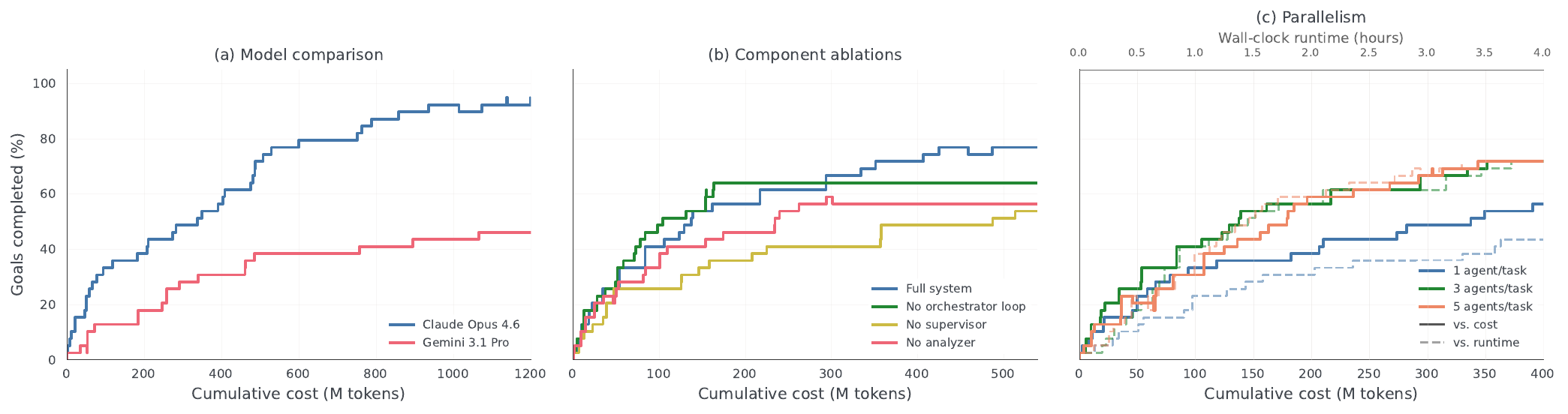}
  \caption{%
    Ablation results on \emph{Algebraic Combinatorics} (39~targets).
    All plots show goals completed (\%) as a function of cumulative
    token cost.
    \textbf{(a)}~Claude vs.\ Gemini with the full pipeline.
    \textbf{(b)}~Removing each feedback component independently.
    \textbf{(c)}~Varying workers per task; solid lines show cost,
    dashed lines show wall-clock runtime (top axis).
  }
  \label{fig:ablations}
\end{figure*}

We evaluate the pipeline along three axes: the underlying language model, the contribution of each feedback component, and the degree of worker parallelism.
All experiments target the same book (\emph{Algebraic Combinatorics} by Richard Stanley, 39~targets) since it was among the smallest books and has a moderate difficulty.
We report cumulative compute cost in millions of tokens (see also Appendix~\ref{app:compute_measure}). 

\paragraph{Model comparison.}
We run the full pipeline with a single worker per task using either Claude~Opus~4.6 or Gemini~3.1~Pro as the backbone (Figure~\ref{fig:ablations}).
At 1200M~tokens, Claude has completed 92\% of targets while Gemini reaches only 46\%.
The gap is visible from the start and, since every other component is identical, is attributable to the model's ability to code in Lean.

\paragraph{Component ablations.}
We remove each of the three feedback components independently and compare against the full pipeline (Figure~\ref{fig:ablations}).
Without the orchestrator loop, a static DAG is created at the start and execution is guided solely by the trace analyzer and supervisor, losing the ability to re-plan at the level of the whole run.
Without the supervisor, the orchestrator receives no structured feedback about target-level quality after merges.
Without the trace analyzer, workers lose inter-attempt learning and cannot build on previous failures.
All four configurations use three workers per task and Claude~Opus~4.6.

Within a matched budget of 600M~tokens, the full system reaches 77\%.
The no-orchestrator variant is the most token-efficient early on, outperforming the full system until ${\sim}100$M~tokens, but plateaus at 64\%---indicating the pipeline got stuck without the orchestrator's ability to re-plan around difficult targets.
Removing the supervisor yields 51\%: without post-merge quality checks the system keeps hitting a wall with no clear signal on how to fix target-level problems.
Removing the trace analyzer reaches 57\% but exhausts its compute budget fastest, as workers repeat the same mistakes across attempts.

\paragraph{Parallelism.}
We vary the number of workers racing on each task---1, 3, or 5---to measure the compute--latency tradeoff (Figure~\ref{fig:ablations}).
In wall-clock time (dashed lines, top axis), parallelism provides a clear speedup: at 4~hours the 3- and 5-agent configurations reach ${\sim}62$--68\% while the single-agent configuration is at 44\%.
Interestingly, 3 and 5~agents also achieve higher scores at lower token budgets, suggesting that parallelism on early, easy tasks reduces wasted serial exploration and lets the pipeline advance faster through the dependency graph.

\subsection{Takeaways and insights}
\label{sec:takeaways}

% \paragraph{Difficulty across books.}
% Most books follow a similar trajectory: the pipeline reaches 50--60\%
% of targets quickly, after which progress slows dramatically. The last
% 20--30\% of targets often consumes most of the total budget.
% Stagnation typically occurs at theorems that require Mathlib
% infrastructure not already present in the library. Books whose content
% aligns closely with Mathlib, such as real analysis perform well since the necessary building blocks
% already exist. Books further from Mathlib's coverage, such as elliptic
% curves or geometry of manifolds, perform substantially worse, as
% formalization requires creating significant infrastructure.
% \charles{redundant with 5.1 - shorten considerably}

\paragraph{Failure modes.}
We observe several recurring failure patterns:
\begin{itemize}[nosep,leftmargin=1.5em]
\item \textbf{Frontal assault.} Workers repeatedly attempt the same proof strategy despite prior failures, wasting tokens on identical dead ends.
\item \textbf{Cheating.} Workers find creative ways to circumvent the build system: introducing hidden axioms deep in the codebase, weakening hypotheses, or smuggling \texttt{sorry} behind helper lemmas. 
A particularly disruptive failure mode occurs when workers replace a foundational but complex mathematical object, such as a manifold or a scheme, with an oversimplified formal definition, which can then contaminate many dependent statements. 
 Stricter reviewing curbs this, but workers then hide axioms in increasingly subtle ways, creating an adversarial dynamic that motivates layered review.
\item \textbf{Infrastructure panic.} When a task requires building substantial infrastructure (e.g., differential forms for manifolds), workers often refuse to proceed, arguing that the task would be too difficult.
\item \textbf{Orchestrator fatigue.} Over long runs, the orchestrator's context degrades: it creates coarse-grained tasks with unhelpful prompts, or silently stops dispatching work on difficult targets.
Delegating failure analysis to the trace analyzer (Section~\ref{sec:ablations}) mitigates this.
\end{itemize}

\paragraph{What worked.} Giving the orchestrator progressive-detail tools, a summary view of target completion rates, the ability to drill into individual targets for the latest feedback, and access to task-level error logs, let it make informed re-planning decisions without flooding its context with the full project state at once.
Delegating responsibilities to specialized agents rather than burdening a single orchestrator distributes cognitive load and avoids context degradation over long runs. Worker escalations, particularly decomposition requests, are surprisingly informative: they surface false axioms and suggest decomposition strategies that make otherwise intractable tasks solvable.

%% ============================================================================
\section{Limitations and Conclusion}
\label{sec:discussion}
%% ============================================================================

We introduced \AutoformBot, a multi-agent system for formalizing mathematical textbooks at scale.
By orchestrating hundreds of LLM agents through proven software engineering practices (version control, review, worktrees, and an issue/task tracker), \AutoformBot produces verified Lean libraries from diverse mathematical sources.
We release two artifacts: \AutoformBot (the open-source framework) and \Atlas (verified formal libraries spanning \textbf{26} books).
Our results suggest that formalizing the core textbook infrastructure of modern mathematics is within reach.
This, in turn, opens the door to the automated verification of new mathematical results, and thus to large scale human/machine collaboration.

However, our system relies on frontier LLMs, whose compute costs still remain nontrivial.
Moreover, each textbook was formalized \textit{in isolation}, without the careful planning needed to make it maximally compatible with existing \Mathlib{} infrastructure.
Human involvement currently remains necessary to handle such organisational concerns---selecting and ordering books in a dependency-aware manner, bridging mismatched conventions across sources, etc.
Our next endeavour will be to bring \Atlas to the level of completeness and standardization required to serve as a proper machine-generated extension of \Mathlib{}, a process where humans-in-the-loop will likely play a key role. 

\section*{Broader Impact}

Like much research on large language models, this work may have significant social impact. In particular, it contributes to a line of research that could substantially reshape the mathematical profession. Our hope is that such tools will ultimately change the field for the better, freeing mathematicians to focus more on the creative and exploratory aspects of their work.

\section*{Acknowledgment}
AH is supported by Hi! PARIS and ANR/France 2030 program (ANR-23-IACL-0005). JK and NP thank 
the Simons Foundation for support through the Collaborative Grant “The Physics of Learning and Neural
Computation” as well as support by the NSF through NRT Award 1922658. 
VC and CA thank Adam Kiezun, Robert Rusch and Shuchao Bi for helpful conversations.

%% file: figures/pipeline.tex
\begin{tikzpicture}[
  >=stealth,
  agent/.style={
    rectangle, rounded corners=2pt, draw=black!60, semithick,
    fill=blue!10, minimum height=8mm, minimum width=22mm,
    font=\scriptsize, align=center
  },
  infra/.style={
    rectangle, rounded corners=2pt, draw=black!30,
    fill=black!4, minimum height=8mm, minimum width=22mm,
    font=\scriptsize, align=center
  },
  flow/.style={->, semithick},
  fb/.style={->, thin, dashed, blue!40!black},
  annot/.style={font=\scriptsize, text=black!50, align=center},
  tier/.style={font=\small\sc, text=black!50},
]

%%% ===== PLANNING TIER =====
\node[font=\small\bfseries, text=black!70] (book) at (-2.4, -1.4)
  {Textbook};
\node[agent] (orch) at (0, -1.4) {Orchestrator};
\node[infra] (dag) at (0, 1.4) {Task DAG};

%%% ===== EXECUTION TIER =====
\node[infra] (skills) at (3.8, 2.2) {Skill Guides};
\node[agent] (workers) at (6.5, 1.4)
  {Workers $\times N$\\[-1pt]{\scriptsize\color{black!45}(git worktrees)}};
\node[agent] (ta) at (3.8, 0)
  {Trace\\[-1pt]Analyzer};
\node[agent] (rev) at (6.5, 0) {Reviewers $\times N$};
\node[infra] (mq) at (6.5, -1.4) {Main codebase};

%%% ===== FEEDBACK TIER =====
\node[agent] (sup) at (9.5, -1.4)
  {Supervisor\\[-1pt]{\scriptsize\color{black!45}(eval harness)}};

%%% ===== MAIN FLOW (solid arrows) =====
\draw[flow] (book) -- (orch);
\draw[flow] (orch) -- (dag);
\draw[flow] (dag) -- node[annot, below] {dispatch} (workers);
\draw[flow] (workers) -- (rev);
\draw[flow] (rev) -- node[annot, right] {merging if approved} (mq);
\draw[flow] (mq) -- (sup);

%%% ===== FEEDBACK LOOPS (dashed arrows) =====

% Supervisor → Orchestrator (feedback, bottom bypass)
\draw[fb] (sup.south) -- ++(0, -0.3) -| (orch.south);
\node[annot, right] at (3.6, -1.9) {feedback};

% Supervisor → Task DAG (fix tasks, top bypass)
\draw[fb] (sup.north) -- ++(0, 3.8) -| (dag.north);
\node[annot, right] at (9.6, 1.0) {fix tasks};

% Workers → Trace Analyzer (failed tasks)
\draw[fb] (workers.south west) --
  node[annot, above, sloped] {failed} (ta.north east);

% Reviewers → Trace Analyzer (rejected)
\draw[fb] (rev.south) -- ++(0, -0.3) -| (ta.south);
\node[annot, below] at (4.75, -0.7) {rejected};

% Trace Analyzer → Skill Guides (writes lessons)
\draw[fb] (ta) -- (skills);

% Skill Guides → Workers (workers read before retrying)
\draw[fb] (skills.east) -- (workers.north west);

% Trace Analyzer → Orchestrator (reports for replanning)
\draw[fb] (ta.south west) --
  node[annot, below, sloped] {reports} (orch.north east);

% Trace Analyzer → Task DAG (fix tasks)
\draw[fb] (ta.north west) -- (dag.south east);

%%% ===== TIER SEPARATORS =====
\draw[black!12, thin, densely dashed] (1.4, -2.5) -- (1.4, 3.6);
\draw[black!12, thin, densely dashed] (8.2, -2.5) -- (8.2, 3.6);

%%% ===== TIER LABELS =====
\node[tier] at (0, -2.5) {Planning};
\node[tier] at (4.8, -2.5) {Execution};
\node[tier] at (9.5, -2.5) {Evaluation};

\end{tikzpicture}

%% file: figures/dependency_cone_II.tex
\begin{tikzpicture}[
  xscale=0.85,
  >=stealth,
  root/.style={
    rectangle, rounded corners=3pt, draw=blue!65!black, semithick,
    fill=blue!55, text=white,
    minimum height=7mm, inner xsep=8pt,
    font=\footnotesize\sffamily\bfseries,
  },
  snode/.style={
    rectangle, rounded corners=3pt, draw=blue!30, thin,
    fill=blue!8,
    minimum height=6mm, inner xsep=5pt,
    font=\footnotesize\sffamily,
  },
  fnode/.style={
    rectangle, rounded corners=3pt, draw=green!45!black!15, thin,
    fill=green!12,
    minimum height=6mm, inner xsep=5pt,
    font=\footnotesize\sffamily,
  },
  dep/.style={->, very thin, draw=black!30},
]

\pgfdeclarelayer{bg}
\pgfsetlayers{bg,main}

% ============= NODES =============
% Positions derived from SVG source, y-compressed by 0.65
\node[root]  (rsf)  at (4.75,  0)      {riemannSum\_eq\_fiber};
\node[snode] (tel)  at (6.9, -0.75)    {telescope};
\node[fnode] (rs)   at (3.3, -1.50)    {riemannSum};
\node[fnode] (asgn) at (8.0, -1.50)    {assign};
\node[fnode] (tags) at (4.4, -2.24)    {tags};
\node[snode] (itr)  at (9.3, -2.24)    {IsTaggedRefinement};
\node[fnode] (tp)   at (7.4, -2.99)    {toPartition};
\node[snode] (tprt) at (4.9, -3.74)    {TaggedPartition};
\node[fnode] (pts)  at (1.8, -4.49)    {points};
\node[fnode] (n)    at (0.8, -5.23)    {n};
\node[snode] (prt)  at (10, -5)   {Partition};

% ============= EDGES (behind nodes) =============
% 39 edges extracted from the SVG source.
% Arrows point from dependency to dependent (upward).
\begin{pgfonlayer}{bg}

% --- To riemannSum_eq_fiber (9 deps) ---
\draw[dep] (tel)  to[out=100, in=-50] (rsf);
\draw[dep] (rs)   to[out=60, in=-145] (rsf);
\draw[dep] (asgn) to[out=110, in=-30] (rsf);
\draw[dep] (tags) to[out=75, in=-110] (rsf);
\draw[dep] (itr.north) .. controls +(0, 1.0) and +(2.5, -0.3) .. (rsf.east);
\draw[dep] (tp)   to[out=90, in=-70] (rsf);
\draw[dep] (tprt.west) .. controls +(-1.0, 0.8) and +(-1.0, -0.8) .. (rsf.west);
\draw[dep] (pts.west)  .. controls +(-1.3, 1.1) and +(-1.3, -1.1) .. (rsf.west);
\draw[dep] (n.west)    .. controls +(-1.8, 1.4) and +(-1.8, -1.4) .. (rsf.west);

% --- To telescope (6 deps) ---
\draw[dep] (itr)  to[out=150, in=-20] (tel);
\draw[dep] (asgn) to[out=170, in=-5] (tel);
\draw[dep] (tp)   to[out=85, in=-90] (tel);
\draw[dep] (tprt.west) .. controls +(-0.8, 0.6) and +(-0.8, -0.6) .. (tel.west);
\draw[dep] (pts.west)  .. controls +(-1.1, 0.9) and +(-1.5, -0.9) .. (tel.west);
\draw[dep] (n.west)    .. controls +(-1.5, 1.2) and +(-2.0, -1.0) .. (tel.west);

% --- To riemannSum (5 deps) ---
\draw[dep] (tags) to[out=150, in=-40] (rs);
\draw[dep] (tp)   to[out=165, in=-20] (rs);
\draw[dep] (tprt) to[out=140, in=-30] (rs);
\draw[dep] (pts)  to[out=100, in=-90] (rs);
\draw[dep] (n.west) .. controls +(-1.0, 0.8) and +(-1.0, -0.8) .. (rs.west);

% --- To assign (4 deps) ---
\draw[dep] (itr)  to[out=120, in=-30] (asgn);
\draw[dep] (tp)   to[out=30, in=-130] (asgn);
\draw[dep] (tprt) to[out=15, in=-155] (asgn);
\draw[dep] (n.east) .. controls +(1.5, 0.9) and +(1.5, -0.9) .. (asgn.east);

% --- To tags (3 deps) ---
\draw[dep] (tprt) to[out=130, in=-50] (tags);
\draw[dep] (tp)   to[out=165, in=-20] (tags);
\draw[dep] (n.west) .. controls +(-0.6, 0.5) and +(-0.6, -0.5) .. (tags.west);

% --- To IsTaggedRefinement (4 deps) ---
\draw[dep] (tprt) to[out=5, in=-170] (itr);
\draw[dep] (tp)   to[out=20, in=-145] (itr);
\draw[dep] (pts.east) .. controls +(1.0, 0.5) and +(1.0, -0.7) .. (itr.east);
\draw[dep] (n.east)   .. controls +(1.5, 1.0) and +(1.5, -1.0) .. (itr.east);

% --- To toPartition (2 deps) ---
\draw[dep] (tprt) to[out=35, in=-145] (tp);
\draw[dep] (prt)  to[out=100, in=-20] (tp);

% --- To TaggedPartition (3 deps) ---
\draw[dep] (pts)  to[out=55, in=-140] (tprt);
\draw[dep] (n)    to[out=60, in=-160] (tprt);
\draw[dep] (prt)  to[out=160, in=-10] (tprt);

% --- To points (2 deps) ---
\draw[dep] (n)    to[out=70, in=-130] (pts);
\draw[dep] (prt.west) to[out=180, in=-10] (pts);

% --- To n (1 dep) ---
\draw[dep] (prt.west) to[out=175, in=-5] (n);

\end{pgfonlayer}

\end{tikzpicture}

%% file: appendix.tex
\section{Measure of compute }
\label{app:compute_measure}

For a given experiment, we estimate the compute spent by summing over:
\begin{itemize}
    \item Regular input tokens, i.e. un-cached tokens that were read while producing a response and are not being committed to the cache,
    \item Cache-read tokens, i.e. cached tokens that were read while producing a response,
    \item Cache-write tokens, i.e. input tokens being committed to the API's cache for the first time,
    \item Output tokens, i.e. tokens produced by the model,
\end{itemize}
with the following coefficients:
\begin{table}[H]
\centering
\renewcommand{\arraystretch}{1.2}
\setlength{\tabcolsep}{10pt}
\begin{tabular}{l c}
\hline
\textbf{Token type (per 1M)} & \textbf{Multiplier} \\
\hline
regular input tokens         & 1x \\
cache-read tokens            & 0.1x \\
cache-write tokens           & 1.25x \\
output tokens                & 5x \\
\hline
\end{tabular}
\end{table}

These coefficients match the typical rates applied by frontier model providers through their APIs, and are a reasonable averaged approximation of the true compute spent.

Moreover, we apply a 0.1 multiplicative discount coefficient to the compute spent on the smaller Haiku 4.5 model (mostly used as a document-reading helper), reflecting the difference in size relative to the bigger flagship models.

In the table below, we report the average percentage of total compute used by each agent type.

\begin{tabular}{lc}
\toprule
\textbf{Category} & \textbf{Ratio (\%)} \\
\midrule
Workers      & $76.35 \pm 5.71$ \\
Reviewers    & $6.86 \pm 2.38$ \\
Supervisor   & $5.72 \pm 1.54$ \\
Orchestrator & $4.01 \pm 3.46$ \\
Full Eval    & $3.80 \pm 2.34$ \\
Readers      & $2.00 \pm 0.35$ \\
Analyzers    & $1.28 \pm 1.65$ \\
\bottomrule
\end{tabular}

\section{Per-Book Detailed Results}
\label{app:per-book}

For each book, we report:
\begin{itemize}
    \item The name of the course.
    \item The name of its author.
    \item The corresponding URL.
    \item The number of statements (definitions, lemmas, theorems, etc.) in the source material that we are aiming to formalize.
    \item The number of statements that were successfully formalized (see Subsection~\ref{sec:eval-harness}).
    \item The number of pages of the source material.
    \item The number of lines of code of the formalization.
    \item The number of tokens spent on the formalization attempt (see Appendix \ref{app:compute_measure} for the exact metric).
\end{itemize}
\bookcard{Algebra Notes}%
  {Michael Artin \& MIT Students}%
  {https://ocw.mit.edu/courses/res-18-011-algebra-i-student-notes-fall-2021/}%
  {176}%             Target statements
  {340}%            Pages
  {4{,}409}%        Lines of code
  {151 (86\%)}%      Formalized statements
  {25}%              Gaps
  {~1,963}%             Tokens (1M)
  {}

\bookcard{Algebraic Combinatorics}%
  {Richard Stanley}%
  {https://ocw.mit.edu/courses/18-318-topics-in-algebraic-combinatorics-spring-2006/}%
  {39}%             Target statements
  {49}%            Pages
  {9{,}343}%        Lines of code
  {37 (95\%)}%      Formalized statements
  {2}%              Gaps
  {~1,441}%             Tokens (1M)
  {}

\bookcard{Algebraic Geometry I}%
  {Roman Bezrukavnikov}%
  {https://ocw.mit.edu/courses/18-725-algebraic-geometry-fall-2015/}%
  {186}%             Target statements
  {63}%            Pages
  {27{,}393}%        Lines of code
  {112 (60\%)}%      Formalized statements
  {74}%              Gaps
  {~7,629}%             Tokens (1M)
  {}

\bookcard{Algebraic Topology I}%
  {Haynes Miller}%
  {https://ocw.mit.edu/courses/18-905-algebraic-topology-i-fall-2016/}%
  {171}%             Target statements
  {109}%            Pages
  {20{,}143}%        Lines of code
  {110 (64\%)}%      Formalized statements
  {61}%              Gaps
  {~10,323}%             Tokens (1M)
  {}

\bookcard{An Algorithmists Toolkit}%
  {Jonathan Kelner}%
  {https://ocw.mit.edu/courses/18-409-topics-in-theoretical-computer-science-an-algorithmists-toolkit-fall-2009/}%
  {158}%             Target statements
  {171}%            Pages
  {8{,}234}%        Lines of code
  {131 (83\%)}%      Formalized statements
  {27}%              Gaps
  {~2,004}%             Tokens (1M)
  {}

\bookcard{Arithmetic Geometry}%
  {Andrew V. Sutherland}%
  {https://ocw.mit.edu/courses/18-782-introduction-to-arithmetic-geometry-fall-2013/}%
  {335}%             Target statements
  {215}%            Pages
  {29{,}573}%        Lines of code
  {266 (79\%)}%      Formalized statements
  {69}%              Gaps
  {~11,101}%             Tokens (1M)
  {}

\bookcard{Boolean Functions}%
  {Dor Minzer}%
  {https://ocw.mit.edu/courses/18-218-topics-in-combinatorics-analysis-of-boolean-functions-spring-2021/}%
  {108}%             Target statements
  {86}%            Pages
  {7{,}949}%        Lines of code
  {44 (41\%)}%      Formalized statements
  {64}%              Gaps
  {~2,327}%             Tokens (1M)
  {}

\bookcard{Buildings}%
  {Paul Garrett}%
  {https://www-users.cse.umn.edu/~garrett/m/buildings/book.pdf}%
  {74}%             Target statements
  {346}%            Pages
  {48{,}809}%        Lines of code
  {44 (59\%)}%      Formalized statements
  {30}%              Gaps
  {~20,443}%             Tokens (1M)
  {}

\bookcard{Combinatorial Optimization}%
  {Santosh Vempala}%
  {https://ocw.mit.edu/courses/18-433-combinatorial-optimization-fall-2003/}%
  {36}%             Target statements
  {66}%            Pages
  {7{,}934}%        Lines of code
  {22 (61\%)}%      Formalized statements
  {14}%              Gaps
  {~2,476}%             Tokens (1M)
  {}

\bookcard{Complex Variables}%
  {Sigurdur Helgason}%
  {https://ocw.mit.edu/courses/18-112-functions-of-a-complex-variable-fall-2008/}%
  {38}%             Target statements
  {100}%            Pages
  {6{,}225}%        Lines of code
  {37 (97\%)}%      Formalized statements
  {1}%              Gaps
  {~1,251}%             Tokens (1M)
  {}

\bookcard{Differential Analysis}%
  {Richard Melrose}%
  {https://ocw.mit.edu/courses/18-155-differential-analysis-fall-2004/}%
  {113}%             Target statements
  {134}%            Pages
  {23{,}713}%        Lines of code
  {88 (78\%)}%      Formalized statements
  {25}%              Gaps
  {~11,743}%             Tokens (1M)
  {}

\bookcard{Differential Geometry}%
  {Paul Seidel}%
  {https://ocw.mit.edu/courses/18-950-differential-geometry-fall-2008/}%
  {147}%             Target statements
  {60}%            Pages
  {8{,}942}%        Lines of code
  {112 (76\%)}%      Formalized statements
  {35}%              Gaps
  {~1,934}%             Tokens (1M)
  {}

\bookcard{Elliptic Curves}%
  {Andrew V. Sutherland}%
  {https://ocw.mit.edu/courses/18-783-elliptic-curves-spring-2021/}%
  {360}%             Target statements
  {306}%            Pages
  {22{,}316}%        Lines of code
  {212 (59\%)}%      Formalized statements
  {148}%              Gaps
  {~11,058}%             Tokens (1M)
  {}

\bookcard{Fourier Analysis}%
  {David Jerison}%
  {https://ocw.mit.edu/courses/18-103-fourier-analysis-fall-2013/}%
  {38}%             Target statements
  {71}%            Pages
  {6{,}671}%        Lines of code
  {34 (89\%)}%      Formalized statements
  {4}%              Gaps
  {~1,186}%             Tokens (1M)
  {}

\bookcard{Geometry Of Manifolds}%
  {Denis Auroux}%
  {https://ocw.mit.edu/courses/18-966-geometry-of-manifolds-spring-2007/}%
  {72}%             Target statements
  {51}%            Pages
  {16{,}408}%        Lines of code
  {40 (56\%)}%      Formalized statements
  {32}%              Gaps
  {~6,865}%             Tokens (1M)
  {}

\bookcard{High Dimensional Statistics}%
  {Philippe Rigollet}%
  {https://ocw.mit.edu/courses/18-s997-high-dimensional-statistics-spring-2015/}%
  {73}%             Target statements
  {131}%            Pages
  {31{,}715}%        Lines of code
  {65 (89\%)}%      Formalized statements
  {8}%              Gaps
  {~975}%             Tokens (1M)
  {}

\bookcard{Introduction To Functional Analysis}%
  {Casey Rodriguez}%
  {https://ocw.mit.edu/courses/18-102-introduction-to-functional-analysis-spring-2021/}%
  {72}%             Target statements
  {176}%            Pages
  {2{,}006}%        Lines of code
  {68 (94\%)}%      Formalized statements
  {4}%              Gaps
  {~554}%             Tokens (1M)
  {}

\bookcard{Introduction To Partial Differential Equations}%
  {Jared Speck}%
  {https://ocw.mit.edu/courses/18-152-introduction-to-partial-differential-equations-fall-2011/}%
  {105}%             Target statements
  {136}%            Pages
  {20{,}740}%        Lines of code
  {86 (82\%)}%      Formalized statements
  {19}%              Gaps
  {~2,972}%             Tokens (1M)
  {}

\bookcard{Representations of Lie Groups}%
  {Pavel Etingof}%
  {https://ocw.mit.edu/courses/18-757-representations-of-lie-groups-fall-2023/}%
  {185}%             Target statements
  {161}%            Pages
  {50{,}594}%        Lines of code
  {74 (40\%)}%      Formalized statements
  {111}%              Gaps
  {~45,384}%             Tokens (1M)
  {}

\bookcard{Number Theory I}%
  {Andrew V. Sutherland}%
  {https://ocw.mit.edu/courses/18-785-number-theory-i-fall-2021/}%
  {576}%             Target statements
  {318}%            Pages
  {54{,}760}%        Lines of code
  {460 (80\%)}%      Formalized statements
  {116}%              Gaps
  {~15,424}%             Tokens (1M)
  {}

\bookcard{Probabilistic Methods in Combinatorics}%
  {Yufei Zhao}%
  {https://ocw.mit.edu/courses/18-226-probabilistic-methods-in-combinatorics-fall-2022/}%
  {210}%             Target statements
  {215}%            Pages
  {15{,}604}%        Lines of code
  {109 (52\%)}%      Formalized statements
  {101}%              Gaps
  {~2,720}%             Tokens (1M)
  {}

\bookcard{Projection Theory}%
  {Lawrence D Guth}%
  {https://ocw.mit.edu/courses/18-156-projection-theory-spring-2025/}%
  {111}%             Target statements
  {142}%            Pages
  {9{,}672}%        Lines of code
  {73 (66\%)}%      Formalized statements
  {38}%              Gaps
  {~2,678}%             Tokens (1M)
  {}

\bookcard{Real Analysis}%
  {Casey Rodriguez}%
  {https://ocw.mit.edu/courses/18-100a-real-analysis-fall-2020/}%
  {177}%             Target statements
  {92}%            Pages
  {2{,}224}%        Lines of code
  {175 (99\%)}%      Formalized statements
  {2}%              Gaps
  {~586}%             Tokens (1M)
  {}

\bookcard{Tensor Categories}%
  {Pavel Etingof}%
  {https://ocw.mit.edu/courses/18-769-topics-in-lie-theory-tensor-categories-spring-2009/}%
  {229}%             Target statements
  {128}%            Pages
  {29{,}729}%        Lines of code
  {137 (60\%)}%      Formalized statements
  {92}%              Gaps
  {~11,338}%             Tokens (1M)
  {}

\bookcard{Theory Of Computation}%
  {Michael Sipser}%
  {https://ocw.mit.edu/courses/18-404j-theory-of-computation-fall-2020/}%
  {118}%             Target statements
  {${\sim}$250}%            Pages
  {10{,}581}%        Lines of code
  {84 (71\%)}%      Formalized statements
  {34}%              Gaps
  {~3,580}%             Tokens (1M)
  {}

\bookcard{Theory Of Probability}%
  {Scott Sheffield}%
  {https://ocw.mit.edu/courses/18-175-theory-of-probability-spring-2014/}%
  {100}%             Target statements
  {${\sim}$175}%            Pages
  {8{,}231}%        Lines of code
  {84 (84\%)}%      Formalized statements
  {16}%              Gaps
  {~3,201}%             Tokens (1M)
  {}

\section{Example Formalizations}
\label{app:examples}

We present three representative formalizations from \Atlas, drawn from different books and mathematical domains.
For each, we show the original textbook statement, the corresponding Lean~4 formalization, and commentary on the modeling choices and proof strategy.

\subsection*{Example 1: Parseval's equality (Boolean Fourier Analysis)}

\paragraph{Source.} Claim~2.2 from Minzer, \emph{Topics in Combinatorics: Analysis of Boolean Functions} (MIT~18.218).

\paragraph{Textbook statement.}
For any $f, g : \{-1,1\}^n \to \mathbb{R}$:
\begin{enumerate}[leftmargin=*]
  \item \emph{Plancherel's equality}: $\langle f, g \rangle = \sum_{S \subseteq [n]} \widehat{f}(S)\, \widehat{g}(S)$.
  \item \emph{Parseval's equality}: $\|f\|_2^2 = \sum_{S \subseteq [n]} \widehat{f}(S)^2$.
\end{enumerate}

\paragraph{Lean formalization.}
\begin{lstlisting}
noncomputable def chi (S : Finset (Fin n))
    (x : Fin n → ZMod 2) : ℝ :=
  (-1 : ℝ) ^ (S.sum fun i => (x i).val)

noncomputable def innerProd
    (f g : (Fin n → ZMod 2) → ℝ) : ℝ :=
  (Finset.univ.sum fun x => f x * g x) / (2 : ℝ) ^ n

noncomputable def fourierCoeff
    (f : (Fin n → ZMod 2) → ℝ)
    (S : Finset (Fin n)) : ℝ :=
  innerProd f (chi S)

theorem parseval (f : (Fin n → ZMod 2) → ℝ) :
    innerProd f f =
    Finset.univ.sum (fun S => fourierCoeff f S ^ 2) := by
  rw [plancherel]; congr 1; funext S; ring
\end{lstlisting}

\paragraph{Commentary.}
The Boolean hypercube $\{-1,1\}^n$ is modeled as \texttt{Fin n $\to$ ZMod 2}, and the Fourier character $\chi_S(x) = \prod_{i \in S} x_i$ becomes $(-1)^{\sum_{i \in S} x_i}$---mathematically equivalent under the isomorphism $0 \mapsto 1, 1 \mapsto {-}1$, but better aligned with \Mathlib{}'s additive group structure on \texttt{ZMod~2}.
The proof follows the textbook's structure: character orthonormality is established via a bit-flipping involution argument (exactly as in the lectures), then Fourier inversion, Plancherel (by bilinearity and orthonormality), and finally Parseval as the special case $g = f$.
The file is sorry-free (153~lines for definitions and all supporting lemmas).

\subsection*{Example 2: Mills' inequality (High-Dimensional Statistics)}

\paragraph{Source.} Proposition~1.1 from Rigollet, \emph{High-Dimensional Statistics} (MIT~18.657).

\paragraph{Textbook statement.}
Let $X \sim \mathcal{N}(\mu, \sigma^2)$. For any $t > 0$,
$$\mathbb{P}(X - \mu > t) \le \frac{1}{\sqrt{2\pi}} \cdot \frac{e^{-t^2/(2\sigma^2)}}{t}.$$
\emph{Proof.} Reduce to $Z \sim \mathcal{N}(0,1)$. For $x \ge t > 0$ we have $1 \le x/t$, so
$\int_t^\infty e^{-x^2/2}\,dx \le \int_t^\infty \tfrac{x}{t}\, e^{-x^2/2}\,dx = \tfrac{1}{t}\, e^{-t^2/2}.$

\paragraph{Lean formalization.}
\begin{lstlisting}
theorem proposition_1_1_mills_inequality
    (t : ℝ) (ht : 0 < t) :
    ∫ x in Ioi t, exp (-(x ^ 2 / 2)) ≤
    t⁻¹ * exp (-(t ^ 2 / 2)) := by
  -- Pointwise: exp(-x^2/2) ≤ (x/t) * exp(-x^2/2)
  have hpointwise : ∀ x ∈ Ioi t,
      exp (-(x ^ 2 / 2)) ≤
      t⁻¹ * (x * exp (-(x ^ 2 / 2))) := ...
  -- Integrate and evaluate RHS via FTC
  calc ∫ x in Ioi t, exp (-(x ^ 2 / 2))
      ≤ ∫ x in Ioi t, t⁻¹ * (x * exp (-(x ^ 2 / 2)))
          := setIntegral_mono_on ...
    _ = t⁻¹ * ∫ x in Ioi t, x * exp (-(x ^ 2 / 2))
          := integral_const_mul _ _
    _ = t⁻¹ * exp (-(t ^ 2 / 2))
          := by rw [integral_Ioi_x_mul_exp t ht]
\end{lstlisting}

\paragraph{Commentary.}
The formalization reduces the general $\mathcal{N}(\mu, \sigma^2)$ statement to the core integral bound for the standard Gaussian, exactly as the textbook proof does.
The key helper lemma \texttt{integral\_Ioi\_x\_mul\_exp} computes $\int_t^\infty x\, e^{-x^2/2}\,dx = e^{-t^2/2}$ using \Mathlib{}'s Fundamental Theorem of Calculus, applied to the antiderivative $f(x) = -e^{-x^2/2}$ whose derivative is $f'(x) = x\, e^{-x^2/2}$.
Separate lemmas establish the derivative via \texttt{HasDerivAt} and the limit $f(x) \to 0$ as $x \to \infty$ via \texttt{tendsto\_exp\_atBot}.
The formalization is sorry-free (130~lines).

\subsection*{Example 3: Sperner's theorem (Algebraic Combinatorics)}

\paragraph{Source.} Corollary~4.8 from Stanley, \emph{Topics in Algebraic Combinatorics} (MIT~18.312).

\paragraph{Textbook statement.}
\emph{Definition~4.2.} A graded poset $P$ of rank $n$ has the \emph{Sperner property} if the maximum antichain size equals $\max_i |P_i|$.

\emph{Corollary~4.8.} The Boolean algebra $B_n$ has the Sperner property.
Equivalently, every antichain in $2^{[n]}$ has at most $\binom{n}{\lfloor n/2\rfloor}$ elements.

\paragraph{Lean formalization.}
\begin{lstlisting}
def HasSpernerProperty (P : GradedPoset α) : Prop :=
  P.maxAntichainCard = P.maxLevelCard

theorem sperner_property_Bn (n : ℕ) :
    (booleanAlgebraGradedPoset n).HasSpernerProperty := ...

theorem sperner_theorem (n : ℕ) :
    ∀ (A : Finset (Finset (Fin n))),
    IsAntichain (· ⊆ ·) (A : Set (Finset (Fin n))) →
    A.card ≤ Nat.choose n (n / 2) := ...
\end{lstlisting}

\paragraph{Commentary.}
The Boolean algebra $B_n$ is modeled as \texttt{Finset (Fin n)} with the subset ordering, inheriting \Mathlib{}'s \texttt{PartialOrder}, \texttt{Fintype}, and \texttt{GradeMinOrder}~$\mathbb{N}$ instances.
The proof follows the textbook's linear-algebraic pipeline over three steps:
(1)~Theorem~4.7 establishes that the \emph{up operator}
$U_i f(T) = \sum_{a \in T} f(T \setminus \{a\})$
is injective on level-$i$ functions when $2i < n$ and surjective when $2i \ge n$, using the commutation relation $D_{i+1}U_i - U_{i-1}D_i = (n-2i)\,I_i$ and an energy/inner-product argument;
(2)~Lemma~4.5 converts these into order-matchings via Hall's marriage theorem, applied to the images of standard basis vectors under the up operator (or its transpose);
(3)~Proposition~4.4 deduces the Sperner property from the existence of upward and downward matchings by constructing a map from any antichain into the peak level, shown to be injective by case analysis on the relative grades.
The concrete form additionally invokes \Mathlib{}'s LYM inequality.
The file is sorry-free (1643~lines).

\section{Evaluation rubrics}
\label{app:rubrics}

Each successfully matched target is scored by three independent LLM judges on a 0--5 integer scale, one judge per rubric.
The judges receive the original book statement, the matched Lean declaration with its source code, and tool access to the declaration dependency graph (Section~\ref{subsec:app_architecture}), which lets them trace sorry chains, inspect structural tags, and investigate suspicious dependencies.
Each judge returns a JSON object containing a numeric score and free-text reasoning.

A target passes only if \emph{every} rubric individually meets its threshold.
The full scoring criteria for each rubric are given below.

\subsection{Faithfulness}
\label{app:rubric-faithfulness}

\paragraph{Purpose.} Evaluates whether the Lean statement faithfully represents the book's mathematical content.

\paragraph{Threshold.} Pass is $\geq 3/5$.

\paragraph{Scoring criteria.}

\begin{description}[leftmargin=1.5em, labelindent=0em]
  \item[5 ---] Captures the same quantifiers, hypotheses, and conclusion structure as the text; preserves local vs.\ global scope; domain conditions and set-theoretic structure match the text and are usable.
  \item[4 ---] Very close to the text: correct locality, correct hypotheses, correct domain conditions; any extra assumptions do not change the mathematical setting (e.g., adding decidability is fine; changing the function space is not).
  \item[3 ---] Core quantifiers, hypotheses, and conclusions match the text; discrepancies are genuinely minor---naming differences, implicit coercions, slightly stronger typeclass assumptions, or equivalent reformulations. The mathematical objects, domain, and all conclusions of the statement are preserved.
  \item[2 ---] Some structure matches, but at least one critical mismatch: wrong underlying type or function space (e.g., \texttt{BoundedContinuousFunction} vs.\ $C_0$), missing a conclusion from a multi-part statement (e.g., missing uniqueness or uniform convergence), adding or dropping hypotheses that change the mathematical content, or formalizing over a different domain than the book specifies.
  \item[1 ---] Major deviations: wrong quantifiers, missing key hypotheses, or incorrect conclusion structure; strengthens or weakens the statement substantially.
  \item[0 ---] Unrelated to the original statement.
\end{description}

\paragraph{Scoring discipline.}
The faithfulness prompt includes an explicit anti-inflation guard.
If the judge's own reasoning describes any discrepancy as ``meaningful'', ``significant'', ``non-trivial'', or ``notable'', the score must be $\leq 2$, since those words are incompatible with the ``genuinely minor'' threshold of score~3.
Specifically: a wrong underlying type or function space is a domain mismatch (score $\leq 2$); a missing conclusion from a multi-part statement is not cosmetic (score $\leq 2$).
Score~3 is reserved for cases where all mathematical objects, domains, and conclusions are correct, and only superficial details differ.

\subsection{Proof integrity}
\label{app:rubric-correctness}

\paragraph{Purpose.} Evaluates mathematical correctness of the formalization---whether hypotheses are appropriate and the Lean statement expresses the intended mathematics rather than a lookalike.

\paragraph{Threshold.} Pass is $\geq 3/5$.

\paragraph{Scoring criteria.}

\begin{description}[leftmargin=1.5em, labelindent=0em]
  \item[5 ---] Accurate encoding of all properties; hypotheses no stronger than needed for the informal statement; the Lean statements express the intended mathematics, not a lookalike.
  \item[4 ---] Accurate encoding of all core properties; placeholders used only where formalization is genuinely heavy, with correct dependencies; assumptions close to minimal.
  \item[3 ---] Key concepts use appropriate Mathlib notions; any gaps are confined to explicitly marked placeholders without changing logical shape; minor definitional mismatches do not undermine intended meaning.
  \item[2 ---] At least one central definition or statement is materially wrong or too weak; properties encoded in a way that does not imply the intended property.
  \item[1 ---] Key notions encoded incorrectly; statements become false or trivial due to vacuous domains, wrong fields, or meaningless placeholders.
  \item[0 ---] Completely wrong or trivially true formalization.
\end{description}

\paragraph{Judge focus.}
The judge is directed to check whether hypotheses are correct and no stronger than needed, whether the conclusion is correct, whether quantifiers, types, and structures match the intended mathematics, and whether the statement could be vacuously true or trivially satisfied due to wrong domains or fields.

\subsection{Code quality}
\label{app:rubric-style}

\paragraph{Purpose.} Evaluates adherence to Lean and Mathlib coding conventions.

\paragraph{Threshold.} Pass is $\geq 3/5$.

\paragraph{Scoring criteria.}

\begin{description}[leftmargin=1.5em, labelindent=0em]
  \item[5 ---] Follows naming convention mapping textbook items to Lean names (e.g., \texttt{theorem\_1\_17}, \texttt{lemma\_1\_5}, \texttt{def\_40\_3}); comments and docstrings indicate correspondence without polluting Lean syntax; imports minimal and relevant; uses existing Mathlib definitions when available.
  \item[4 ---] Clear structure and naming; most declarations correspond cleanly to textbook items; comments and docstrings are informative; imports are largely appropriate; new definitions introduced only when justified.
  \item[3 ---] Most items can be traced back to the text; naming mostly consistent; basic organization present but could be clearer; some redundant imports or non-idiomatic redefinitions.
  \item[2 ---] Partial mapping to text but inconsistent naming conventions; comments or docstrings missing or unhelpful; imports heavy or redundant; ignores available Mathlib primitives.
  \item[1 ---] Naming and structure are chaotic; cannot map Lean items to textbook statements; comments pollute Lean syntax; over-imports or redefines basic Mathlib concepts.
  \item[0 ---] Unreadable or structurally broken.
\end{description}

\paragraph{Judge focus.}
The judge is directed to assess whether each Lean item can be traced back to a textbook item, whether naming follows the project's conventions, whether comments and docstrings are informative without polluting Lean syntax, whether imports are minimal, and whether the formalization reuses Mathlib definitions when available.

\subsection{Score aggregation}

%Individual rubric scores are normalized to $[0,1]$ by dividing by the maximum score (5).
%The overall statement score is the weighted average:
%\[
%  s = 0.4 \cdot s_{\text{faith}} + 0.4 \cdot s_{\text{integrity}} + 0.2 \cdot s_{\text{quality}}.
%\]
A target passes only if every individual rubric meets its threshold ($\geq 3/5$), which ensures a satisfactory degree of completeness. 
%A high weighted average with one rubric below threshold still results in a failure, enforcing minimum quality across all dimensions.

\subsection{Estimation of containment within \Mathlib{}}
\label{app:containment_rubric}
\paragraph{Scoring criteria}

\begin{description}[leftmargin=1.5em, labelindent=0em]
  \item[5 ---] \textbf{Fully contained.} The exact statement (or a direct, trivially equivalent reformulation) exists in mathlib as a named declaration. A user could cite it directly, or with a thin wrapper.
  \item[4 ---] \textbf{Substantially contained.} The statement is provable in a few lines from existing mathlib lemmas, or a strictly more general version is in mathlib and the target is an immediate specialization.
  \item[3 ---] \textbf{Partially contained.}  Core ingredients (definitions and key supporting lemmas) exist in mathlib, but the headline statement itself is not there and would require nontrivial assembly.
  \item[2 ---] \textbf{Minimally contained.} Only background definitions or unrelated prerequisites are available; the substantive content is missing.
  \item[1 ---] \textbf{Not contained.} Mathlib does not contain the statement or the
  necessary specialized definitions.
\end{description}

\section{Agent Prompts and Role Descriptions}
\label{app:prompts}

Each agent in the pipeline is defined declaratively by a system prompt (\texttt{prompt.md}) and a configuration file (\texttt{config.yaml}) specifying its model, turn budget, tool timeout, and the set of MCP tool servers it may access.
All agents use Claude Opus~4.6 unless otherwise noted.
Below we summarize the role, tooling, and key instructions for each agent type.

\paragraph{Orchestrator.}
The orchestrator is a long-lived planning agent (up to 100{,}000 turns, 400K context window with compaction at 70\% utilization) with read-only filesystem access, DAG-store tools, and a persistent TODO list.
It never writes Lean code.
Its prompt describes the workspace layout (\texttt{book/}, \texttt{code/}, \texttt{skills/}, \texttt{reports/}), the full DAG tool API (\texttt{list\_items}, \texttt{add\_item}, \texttt{update\_item}, \texttt{delete\_item}, \texttt{dispatch\_task}, \texttt{dispatch\_ready}), a goal tracker for target-level status, git inspection tools, and an escalation channel.

The prompt specifies a strict task granularity rule: each task covers at most one mathematical statement or one specific fix (a single \texttt{sorry}, a single axiom, a single faithfulness issue).
It details the first-round workflow (read the book, create one task per formalizable statement with dependency edges from the book's logical structure, maximize parallelism within each dependency layer) and subsequent-round workflows (read reports from the trace analyzer, update or split failed tasks, cross-reference against goal status).
The orchestrator is instructed to prioritize failed goals over pending ones, never silently drop scope, and never retry a failed task with an identical prompt.
It also carries the full anti-cheating taxonomy (see below).

\paragraph{Worker.}
The worker is a short-lived formalization agent (up to 250 turns, 300s tool timeout) with read-write filesystem access, a Lean REPL for interactive type-checking, a Lean LSP server for in-file diagnostics, \Mathlib{} search tools (Loogle-based), and git.
Each worker operates in an isolated git worktree branched from \texttt{main}.

The prompt describes the workspace layout and instructs the worker to read skill guides in a specific order before writing any code: first, a task-specific guide (\texttt{skills/tasks/\textlangle{}id\textrangle{}/guide.md}) containing lessons from previous failed attempts at the same task; then Lean/\Mathlib{} API reference files; then workflow best-practice guides.
Workers are instructed to use absolute paths, commit with the task ID as prefix, keep helper lemmas public (not \texttt{private}), and use \texttt{read\_and\_summarize} for large files to preserve context.
The prompt includes an escalation protocol: workers may call \texttt{escalate(severity, message)} only for infrastructure failures or tool malfunctions, never for difficult proofs or slow progress.

The prompt includes a detailed ``no cheating'' section (reproduced in full below), which is shared across the worker, reviewer, and orchestrator prompts.

\paragraph{Reviewer.}
The reviewer is a code-review agent (up to 40 turns, 120s tool timeout) with a Lean LSP server, read-only filesystem access, and git---but no Lean REPL or \Mathlib{} search.
It evaluates each worker's changes against five criteria:
\begin{enumerate}[nosep]
  \item \textbf{Compilation}: run \texttt{lean\_diagnostic\_messages} on changed files.
  \item \textbf{Task completion}: compare the formalization against the original LaTeX source in \texttt{book/}---the reviewer is explicitly instructed to read the book directly and \emph{not} rely on the worker's docstrings or notes.
  \item \textbf{Mathematical correctness}: check proof logic and definitions.
  \item \textbf{Conventions}: proper imports, naming, and structure.
  \item \textbf{Anti-cheating}: actively check for dishonest proof techniques (see below).
\end{enumerate}
The reviewer is specifically instructed that extra hypotheses not present in the book are ``deviations, not justifications''---if the book does not assume a condition, the reviewer must reject unless the worker proves it follows from the book's assumptions.
Responses are structured as either \texttt{APPROVED: \textlangle{}reason\textrangle{}} or \texttt{REJECTED: \textlangle{}specific feedback\textrangle{}} with file paths, line numbers, and concrete fix suggestions.

\paragraph{Trace analyzer.}
The trace analyzer is a persistent agent (up to 100{,}000 turns) assigned to each failed task, retaining full conversation history across all attempts.
It has read-only filesystem access and a trace inspector providing access to execution traces: build errors, reviewer feedback, tool call sequences, agent reasoning, and worker escalations.

The prompt specifies exactly three outputs:
\begin{enumerate}[nosep]
  \item A JSON report (\texttt{reports/\textlangle{}id\textrangle{}.json}) with task ID, status, attempt count, a 1--2 sentence summary, and up to 3 suggestions.
  \item A task-specific skill guide (\texttt{skills/tasks/\textlangle{}id\textrangle{}/guide.md}), written only on failure, containing the exact code that almost worked, correct \Mathlib{} API names, proof strategies to try (and which ones failed), and specific error messages with fixes.
  \item An escalation recommendation when the problem is beyond the pipeline's ability to self-correct.
\end{enumerate}
The prompt includes a hard rule: never suggest \texttt{sorry} unless the result has no informal proof in the book.

\paragraph{Merge matcher.}
The merge matcher (up to 1{,}000 turns) has read-only filesystem access (restricted to \texttt{read\_text\_file}, \texttt{file\_grep}, \texttt{search\_files}, \texttt{list\_directory}).
Given a git diff from a recent merge and a numbered list of book targets, it inspects the changed files and book source to determine which targets are affected.
It returns a JSON response with the list of affected target indices and reasoning.

\paragraph{Matcher (evaluation harness).}
The matcher agent (up to 1{,}000 turns) has read-only filesystem access.
Given an informal mathematical statement from the textbook, it searches the Lean source directory to find the corresponding declaration.
The prompt specifies a search protocol: list the directory structure, use \texttt{file\_grep} with regex patterns to find declarations, read surrounding context with offset/limit, and check \texttt{namespace} blocks to determine fully qualified names.
For multi-part statements, the matcher selects the strongest single declaration and notes related ones.
Responses include the declaration name, file path, a confidence level (\texttt{high}, \texttt{medium}, \texttt{low}, or \texttt{not\_found}), and reasoning.

\paragraph{Judge (evaluation harness).}
The judge agent (up to 1{,}000 turns) has access to \Mathlib{} search tools (\texttt{mathlib\_grep}, \texttt{mathlib\_find\_name}, \texttt{mathlib\_read\_file}) and read-only filesystem access.
Given a mathematical statement and its Lean formalization, it scores the formalization on a 0--5 scale according to a specific rubric (faithfulness, proof integrity, or code quality).
Each judge invocation receives the rubric criteria, scoring guidelines, and the rubric-specific prompt template as part of its input.
The response is a JSON object with a numeric score and reasoning.

\paragraph{Anti-cheating taxonomy.}
The worker, reviewer, and orchestrator prompts all share a detailed taxonomy of dishonest formalization patterns, reproduced here in full:
\begin{enumerate}[label=(\alph*),nosep,leftmargin=*]
  \item \textbf{Trivial statement substitution.}
    Replacing a theorem's statement with \texttt{True} or another trivially provable proposition while keeping its name and docstring (e.g., \texttt{theorem bezout\_theorem : True := by trivial}).
  \item \textbf{Encoding theorems as definitions.}
    Writing \texttt{def foo (...) : Prop := \textlangle{}statement\textrangle{}} for something the textbook presents as a theorem.
    The definition always type-checks (a \texttt{Prop} is just a type), so nothing is proved.
  \item \textbf{Smuggling assumptions into structure fields.}
    Defining a structure whose fields include what should be proved as theorems, then deriving consequences ``for free.''
    Anything stated by the textbook as a ``Theorem,'' ``Proposition,'' ``Corollary,'' or ``Lemma'' must be a separate Lean theorem proved from the class fields---never a class field itself.
  \item \textbf{Weakening the mathematical content.}
    Proving a weaker or purely numerical shadow of a theorem instead of the actual result---for instance, proving two vector spaces have the same dimension instead of constructing an isomorphism, or proving a result about integers that encodes a geometric theorem without ever constructing the geometric objects.
  \item \textbf{Modeling avoidance.}
    Replacing the mathematical objects the textbook works with (e.g., manifolds, schemes, group representations) by simpler algebraic proxies without proving the proxy faithfully represents the real object.
  \item \textbf{Unacknowledged sorry/axiom.}
    Using \texttt{sorry} or \texttt{axiom} in helper lemmas that are then called by ``proved'' theorems, so the top-level theorem appears complete but rests on unproved foundations.
\end{enumerate}

\section{Structural Tags in the Dependency Graph}
\label{subsec:app_architecture}

As described in Section~\ref{sec:eval-harness}, the evaluation harness builds a declaration-level dependency graph and computes \emph{structural tags} for each declaration by pattern-matching on the proof term.
Tags found anywhere in a declaration's dependency cone propagate upward as \emph{alerts}.
Ten tags are currently detected, each flagging a pattern that \emph{may} indicate a formalization defect---though some have legitimate uses:

\begin{itemize}[leftmargin=*]
  \item \textbf{\texttt{vacuous\_body}.}
    The body reduces to \texttt{True}, \texttt{PUnit.unit}, \texttt{trivial}, or a literal.
    Flags theorems whose statement may have been silently replaced by a tautology.

  \item \textbf{\texttt{ignores\_params}.}
    A lambda abstraction that never references its bound variables.
    Flags definitions or theorems that accept parameters for type-checking purposes but discard them, often indicating a stub.

  \item \textbf{\texttt{proof\_by\_exfalso}.}
    The proof proceeds via \texttt{False.elim}, \texttt{absurd}, or similar.
    A valid technique when the context is genuinely contradictory, but can also indicate that an unsatisfiable hypothesis was smuggled into the statement.

  \item \textbf{\texttt{proof\_by\_subsingleton}.}
    The proof invokes \texttt{Subsingleton.elim}.
    Legitimate when the target type is provably a subsingleton, but can mask a theorem whose conclusion has been trivialized.

  \item \textbf{\texttt{returns\_assumption}.}
    The body is a single bound variable, or a structure-field projection applied to a bound variable---i.e., the ``proof'' simply returns one of its hypotheses.
    Flags theorems where what the textbook states as a result has been encoded as an assumption.

  \item \textbf{\texttt{field\_projection\_body}.}
    The body extracts a field from a structure rather than proving a substantive result.
    Flags declarations where a theorem's content may have been packed into a structure definition and then projected out.

  \item \textbf{\texttt{custom\_hypothesis\_in\_type}.}
    The type contains an instance-implicit argument (\texttt{[inst~:~MyClass~X]}) for a project-defined class.
    Flags potential assumption smuggling: the class may bundle axioms that should instead be proved as separate theorems.

  \item \textbf{\texttt{trivial\_constructor}.}
    The body applies a constructor of a project-defined type, but none of the constructor arguments reference project-local declarations.
    Flags instances assembled from library defaults rather than genuine mathematical content.

  \item \textbf{\texttt{orphan\_class}.}
    A project-defined class with zero instances anywhere in the project.
    Flags classes likely introduced to bundle assumptions without ever being instantiated by concrete mathematical objects.

  \item \textbf{\texttt{trivial\_instance}.}
    An instance declaration for a project-defined class whose body carries one of the suspicious tags above (e.g. \texttt{vacuous\_body}, \texttt{ignores\_params}).
    Flags instances that satisfy a typeclass vacuously rather than by genuine construction.
\end{itemize}

The dependency graph serves two roles in the evaluation.
First, it enables \emph{inherited-failure filtering}: when a declaration fails the axiom-purity check solely because of \texttt{sorryAx} inherited from a transitive dependency that is itself an evaluated target with the same violation, the failure is attributed to the upstream root cause and the downstream declaration is not penalized.
This prevents a single \texttt{sorry} from cascading into spurious failures across all its dependents.
Second, the graph is exposed to the LLM judges as a set of queryable tools---including dependency-health summaries, sorry-chain tracing, and suspicious-node listing---allowing them to investigate structural patterns when scoring faithfulness and proof integrity.

\section{Human Expert Assessment}
\label{app:annotators_feedback}

We asked professional mathematicians with Lean expertise to independently review the machine-generated formalization of \emph{Topics in Algebraic Combinatorics} by R.~Stanley, one of our reference textbooks. Below is the full, unedited feedback we received.

\medskip

\paragraph{Overall verdict.}
The formalization does not compile in Lean~4.30, but instead targets Lean~4.28.
It also uses two explicit axioms---\texttt{youngAdjMatrix\_eigenvalues\_bridge} and \texttt{spectral\_trace\_pow}---which are placeholders for results that are not formalized.

\paragraph{Circulant Hadamard section.}

\begin{itemize}[leftmargin=*]
  \item \textbf{Theorem~1} (no circulant Hadamard matrix of order $2^k$ for $k\ge 3$): Ok.
  \item \textbf{Lemma~2} ($x^{2^{k-1}}+1$ irreducible over $\mathbb{Q}$): Ok.
  \item \textbf{Unique expression in the $\zeta$-basis}: Ok.
  \item \textbf{Circulant eigenvalue/determinant formula}: Ok.
  \item \textbf{Lemma~3} (Fourier eigenvalues have absolute value $\sqrt{n}$; factorization of~$2$ in $\mathbb{Z}[\zeta]$): Ok.
  \item \textbf{Lemma~5} ($\mathbb{Z}[\zeta]/(1-\zeta)\cong\mathbb{F}_2$): Ok.
  \item \textbf{Lemma~6}: Phrasing varies from the book.
    The formalization gives equation~(3) as a hypothesis, when the formula should be proved as a lemma.
  \item \textbf{Corollary~7}: The book's statement concerns two specific eigenvalues of~$H$; the formalization states a more general fact, changing the content of the lemma.
  \item \textbf{Lemma~8} (algebraic integer with all conjugates of absolute value one is a root of unity): Ok.
  \item \textbf{Theorem~9 (Kronecker)}: The formalization corrects the book's statement by adding the hypothesis that $\alpha$ is integral over~$\mathbb{Z}$---this is mathematically necessary, as the original statement is false (e.g.\ $(3+4i)/5\in\mathbb{Q}(i)$ has absolute value one but is not a root of unity). The Lean statement is also slightly more general, allowing the field extension to be generated by multiple roots of unity.
  \item \textbf{Final coefficient comparison (proof of Theorem~1)}: Ok; the formalization also fixes a typo in the book.
\end{itemize}

\paragraph{Section~4: The Sperner property.}

\begin{itemize}[leftmargin=*]
  \item \textbf{Definition~4.1}: Ok.
    Essentially rephrases Mathlib's built-in \texttt{GradeMinOrder}.
  \item \textbf{Definition~4.2} (Sperner property): Ok.
  \item \textbf{Order-matching definition}: Ok.
  \item \textbf{Proposition~4.4} (order-matchings imply rank-unimodal and Sperner): Ok.
  \item \textbf{Lemma~4.5} (injective/surjective order-raising linear operators give matchings): Ok.
  \item \textbf{Boolean up and down operators}: Ok.
  \item \textbf{Lemma~4.6} ($D_{i+1}U_i - U_{i-1}D_i = (n-2i)I_i$): Ok.
  \item \textbf{Theorem~4.7} (Boolean up operator injectivity/surjectivity): Ok.
  \item \textbf{Corollary~4.8} ($B_n$ is Sperner): Ok.
\end{itemize}

\paragraph{Section~5: Group actions on Boolean algebras.}

\begin{itemize}[leftmargin=*]
  \item \textbf{Definition~5.4} (quotient poset $B_n/G$): Ok.
  \item \textbf{Proposition~5.6} ($B_n/G$ graded of rank~$n$ and rank-symmetric): The Lean file defines ``graded'' using an axiomatic rank function rather than the book's definition in terms of maximal chains.
  \item \textbf{Lemma~5.7} (orbit sums form a basis for the fixed subspace): Ok.
  \item \textbf{Lemma~5.8} (Boolean up operator preserves the $G$-fixed subspace): Ok.
  \item \textbf{Theorem~5.9} ($B_n/G$ is graded, rank-symmetric, rank-unimodal, and Sperner): Ok.
  \item \textbf{Theorem~5.10} (nonisomorphic simple graphs; symmetric and unimodal edge counts): Ok.
\end{itemize}

\paragraph{Section~6: Young diagrams and $q$-binomial coefficients.}

\begin{itemize}[leftmargin=*]
  \item \textbf{Partition and Young-lattice setup}: Ok; mathematically equivalent, though stated differently.
  \item \textbf{Proposition~6.2} ($L(m,n)$ graded of rank~$mn$, rank-symmetric): The Lean file states rank-symmetry via the existence of an involution, which is equivalent but differs from the book's definition.
  \item \textbf{Proposition~6.3} ($|L(m,n)|=\binom{m+n}{m}$): The Lean file redefines $L(m,n)$ in a way that simplifies the proof, despite having already defined it differently.
  \item \textbf{$q$-number, $q$-factorial, $q$-binomial definitions}: Done differently than in the book.
    \texttt{qBinomial} is recursive (taking Lemma~6.5 as a definition) rather than a quotient of $q$-factorials; a separate theorem relates the two.
  \item \textbf{Lemma~6.5} ($q$-Pascal recurrence): Ok.
  \item \textbf{Theorem~6.6} (rank generating function of $L(m,n)$): Ok.
  \item \textbf{Definition of $G_{mn}$}: Ok.
  \item \textbf{Lemma~6.8} (every $G_{mn}$-orbit contains exactly one Young diagram): Ok.
  \item \textbf{Theorem~6.9} ($B_{R_{mn}}/G_{mn}\cong L(m,n)$): The first half is stated as a bijection between $L(m,n)$ and Young diagrams but does not mention the quotient by the group action.
  \item \textbf{Corollary~6.10} ($L(m,n)$ is rank-symmetric, rank-unimodal, and Sperner): Ok.
  \item \textbf{Subset-sum setup}: Ok.
  \item \textbf{Theorem~6.11}: Ok, but slightly generalized to a strictly monotone real sequence instead of an arbitrary positive set.
\end{itemize}

\paragraph{Section~8: A glimpse of Young tableaux.}

\begin{itemize}[leftmargin=*]
  \item \textbf{Young tableaux and $f^\lambda$}: Defines \texttt{numSYT} by saturated-chain/operator count, not by an explicit type of standard Young tableaux.
    These are mathematically equivalent, but the tableau object and bijection are not formalized---one of the main nonliteral modelling choices.
  \item \textbf{Hasse-walk count $\alpha(w,\lambda)$ and Equation~(40)}: Ok.
  \item \textbf{Lemma~8.1}: Ok.
  \item \textbf{Valid $\lambda$-word definition}: Ok.
  \item \textbf{Young-lattice up and down operators}: Ok.
  \item \textbf{Lemma~8.2} ($D_{i+1}U_i - U_{i-1}D_i = I_i$): Ok.
  \item \textbf{Equation~(43)} ($DU^i = U^i D + iU^{i-1}$): Ok.
  \item \textbf{Theorem~8.3}: Ok.
  \item \textbf{Corollary~8.4} ($\sum_{\lambda\vdash n}(f^\lambda)^2 = n!$): Ok.
  \item \textbf{$\beta(\ell,\lambda)$ setup and facts F1/F2}: Ok.
  \item \textbf{Lemma~8.5} (normal-order expansion coefficients): Ok.
  \item \textbf{Theorem~8.6}: Ok.
  \item \textbf{Corollary~8.7} (closed Hasse walks of length $2m$ from $\varnothing$): Ok.
  \item \textbf{Theorem~8.8} (eigenvalues of the bipartite graph $Y_{j-1,j}$):
    \textbf{Not ok.}
    The finite-matrix spectral conclusion depends on the explicit axiom \texttt{youngAdjMatrix\_eigenvalues\_bridge}.
    The Lean predicate \texttt{IsEigenvectorYoung} asserts only the eigen-equation without requiring a nonzero vector.
    The construction works with global functions on all partitions rather than fully level-restricted finite spaces.
  \item \textbf{Corollary~8.9} (alternating walk count):
    \textbf{Not ok.}
    Lean proves the combinatorial half-trace part, but the spectral trace identity is supplied by the explicit axiom \texttt{spectral\_trace\_pow}, and the eigenvalue input depends on \texttt{youngAdjMatrix\_eigenvalues\_bridge}.
\end{itemize}

\paragraph{General comments.}
The main issues are the two added axioms in the last two statements.
Overall the project was built in a way that was somewhat inconvenient to use, as the files often imported other files, requiring the project to be set up in a specific way.
There was no note that the formalization did not target the newest version of Lean, which should have been included.
The Lean formalization fixed some small mistakes in the book and skipped the examples.
Overall, most of the formalization is solid, but the hardest statements are not formalized faithfully.